\documentclass{article}

\usepackage[final,nonatbib]{neurips_fitml_2024}
\usepackage[numbers]{natbib}

\usepackage[utf8]{inputenc} % allow utf-8 input
\usepackage[T1]{fontenc}    % use 8-bit T1 fonts
\usepackage{hyperref}       % hyperlinks
\usepackage{url}            % simple URL typesetting
\usepackage{booktabs}       % professional-quality tables
\usepackage{amsfonts}       % blackboard math symbols
\usepackage{nicefrac}       % compact symbols for 1/2, etc.
\usepackage{microtype}      % microtypography
\usepackage{xcolor}         % colors

\usepackage{amsmath,amsfonts,bm}
\usepackage{amsmath}
\usepackage{amssymb}
\usepackage{mathtools}
\usepackage{amsthm}

\theoremstyle{plain}

\theoremstyle{definition}

\theoremstyle{remark}

\newcommand{\alg}{\mathcal{A}}
\newcommand{\unlearn}{\mathcal{U}}

\newcommand{\dataset}{\mathcal{D}}
\newcommand{\forgetset}{\mathcal{D}_f}
\newcommand{\retainset}{\mathcal{D}_r}
\newcommand{\testset}{\mathcal{D}_t}
\newcommand{\refine}{\mathcal{F}}

\usepackage{caption} 
\usepackage{tabularx} 
\usepackage{subcaption} 
\usepackage{threeparttable}
\usepackage{graphicx}
\usepackage{subcaption}

\usepackage{wrapfig,lipsum,booktabs}

\title{Scalability of memorization-based machine unlearning }

\author{%
    Kairan Zhao
    \thanks{Correspondence to Kairan.Zhao@warwick.ac.uk}\\
  University of Warwick\\
  \And
  Peter Triantafillou \\
  University of Warwick \\
}

\begin{document}

\maketitle

\begin{abstract}
Machine unlearning (MUL) focuses on removing the influence of specific subsets of data (such as noisy, poisoned, or privacy-sensitive data) from pretrained models. 
MUL methods typically rely on specialized forms of fine-tuning. 
Recent research has shown that data memorization is a key characteristic defining the difficulty of MUL. 
As a result, novel memorization-based unlearning methods have been developed, demonstrating exceptional performance with respect to unlearning quality, while maintaining high performance for model utility. 
Alas, these methods depend on knowing the memorization scores of data points and computing said scores is a notoriously time-consuming process. This in turn severely limits the scalability of these solutions and their practical impact for real-world applications.  
In this work, we tackle these scalability challenges of state-of-the-art memorization-based MUL algorithms using a series of memorization-score proxies. 
We first analyze the profiles of various proxies and then evaluate the performance of state-of-the-art (memorization-based) MUL algorithms in terms of both accuracy and privacy preservation. Our empirical results show that these proxies can introduce accuracy on par with full memorization-based unlearning while dramatically improving scalability. We view this work as an important step toward scalable and efficient machine unlearning.
\let\thefootnote\relax\footnotetext{Code is available at: \url{https://github.com/kairanzhao/RUM}}
% \footnote{Code is available at: \url{https://github.com/kairanzhao/RUM}}
\end{abstract}

\section{Introduction}
Deep learning models have achieved significant success across various domains, largely by increasing model capacity and utilizing vast amounts of data. However, real-world training data often includes examples that may be polluted, harmful, or privacy-sensitive. This raises the need for methods to remove the influence of such undesirable data from pre-trained models.
To address this challenge, machine unlearning (MUL) was introduced \cite{cao2015towards,nguyen2022survey}. MUL aims to remove the impact of a specific subset of training data, ensuring that a model "forgets" the knowledge derived from it. As concerns over data integrity and privacy continue to grow \cite{rosen2011right, hoofnagle2019european}, MUL has emerged as a rapidly growing research area, gaining significant attention in recent years \cite{triantafillou2024we}.

MUL can be viewed as a form of fine-tuning, which is operated on a pre-trained model to specifically "unlearn" the influence of a selected set of training data examples.
While traditional fine-tuning adjusts a model to improve performance on new tasks or additional data, MUL takes a reverse approach: it modifies the model to eliminate the effects of certain data points, ensuring that they no longer contribute to the model's behavior. This process is crucial in contexts where some data must be forgotten due to ethical, legal, or privacy concerns, or to correct erroneous information.

An important aspect of both fine-tuning and unlearning is the role of memorization in deep learning models. Deep neural networks with sufficient capacity are known to memorize their training data \cite{arpit2017closer}, and recent theoretical work has shown that memorization is crucial for achieving near-optimal generalization, particularly in cases where the training data distribution is long-tailed \cite{feldman2020does}. Memorization allows models to retain rare or atypical examples, which can enhance performance on difficult tasks. 
In the context of MUL, it has long been conjectured that training-example memorization is also a key factor in the MUL process. 
Recently, in large language models (LLMs), new algorithms were presented that exploit the memorization of textual sequences, offering new state-of-the-art MUL performance for memorized-data unlearning \cite{barbulescu2024each}.
% Recently for the first time, 
Furthermore, Zhao et al. \cite{zhao2024makes} showed that memorization is strongly linked to the difficulty of the unlearning task: the more memorized the data is, the harder it is to effectively unlearn those examples.
Building on these findings, Zhao et al. \cite{zhao2024makes} proposed a new meta-algorithm, "RUM," which leverages varying levels of memorization to improve existing approximate unlearning algorithms. However, this approach has practical limitations, as it requires precise knowledge of memorization levels in the dataset, which is computationally expensive to obtain. This severely limits the scalability of high-performing MUL algorithms like RUM.

Motivated by this limitation, we explore and adopt a series of memorization proxies to ensure scalability while maintaining the effectiveness of this new class of high-performing machine unlearning algorithms, such as RUM. 
By using proxies that can be computed more efficiently, we aim to strike a balance between performance and computational feasibility.

\section{Related work and background}
\label{background}

\subsection{Problem Formulation}
Let $\theta_o = \alg(\dataset_{train})$ denote the weights of a deep neural network trained on a dataset $\dataset_{train}$ using the algorithm $\alg$; we refer to $\theta_o$ as the "original model" in an unlearning task. 
Suppose we have a subset $\forgetset \subseteq \dataset_{train}$ that we wish to remove its influence from the model, defined as the "forget set". The complement of this subset, $\retainset = \dataset_{train} \setminus \forgetset$, is referred to as the "retain set", representing the data whose knowledge we aim to preserve.
The unlearning process involves applying an unlearning algorithm $\unlearn$ to the original model, resulting in $\theta_u = \unlearn(\theta_o, \forgetset, \retainset)$. The goal of unlearning is for $\theta_u$ to approximate the model $\theta_r$ that would have been obtained by retraining from scratch solely on $\retainset$.

\subsection{Memorization score and proxies}
\paragraph{Memorization \citep{feldman2020does}}
Memorization, as defined by Feldman \cite{feldman2020does}, measures the extent to which a machine learning model's predictions rely on a specific training data example. An example is considered memorized if the model’s performance changes significantly when the example is included or removed.
Studies have shown that atypical or outlier examples, particularly those with noisy or incorrect labels, are more likely to be memorized \cite{feldman2020does, feldman2020neural, jiang2020characterizing}.

Formally, for a data point $(x_i, y_i) \in \dataset$, where $x_i$ is the feature and $y_i$ is the label, the \textit{memorization score} with respect to a training dataset $\dataset$ and algorithm $\alg$ is given by:

\begin{equation} \text{mem}(\alg, \dataset, i) = \Pr_{f \sim \alg(\dataset)}[f(x_i) = y_i] \ - \Pr_{f \sim \alg(\dataset \setminus i)}[f(x_i) = y_i] \end{equation}

where the first term considers models trained on the entire dataset, while the second reflects models trained without the example $(x_i, y_i)$. 
A high memorization score indicates that excluding the example causes a significant change in the model's predictions for that example.

Although Feldman et al. \cite{feldman2020neural} proposed methods to estimate memorization, these approaches require training numerous models on different dataset splits, making them computationally expensive and impractical for deep learning models.
To address this, researchers have developed several alternative metrics that can act as proxies for memorization, with key examples outlined below:

\paragraph{C-score \cite{jiang2020characterizing}}
% \todo{should we include c-score?}
C-score, introduced by Jiang et al. \cite{jiang2020characterizing}, measures the alignment of a held-out data example with the underlying data distribution \(\mathcal{P}\). For a given example $(x_i, y_i)$, it evaluates the expected performance of models trained on increasingly larger subsets of data sampled from \(\mathcal{P}\), excluding $(x_i, y_i)$. This consistency profile reflects how structurally aligned an example is with the distribution \(\mathcal{P}\).
Notably, a data point evaluation of the consistency profile at a fixed data size resembles the second term of the memorization score formula. 
Since the C-score estimation follows a similar process to  the memorization estimator proposed by Feldman et al. \cite{feldman2020neural}, it remains computationally expensive.

\paragraph{Learning events proxy \cite{toneva2018empirical, jiang2020characterizing}} 
The learning events proxy is introduced by Jiang et al. \cite{jiang2020characterizing}, which is a class of proxies designed to measure how quickly and reliably a model learns a specific example during training. 
For a given data example $(x_i, y_i) \in \dataset$, learning events proxies are computed by collecting several metrics at each training epoch as the model $\theta$ is trained on $\dataset$ using algorithm $\alg$, and averaging these metrics over all epochs. 
The key metrics include: 
\textbf{confidence}, which is the softmax probability of $\theta(x_i)$ corresponding to the ground-truth label $y_i$; 
\textbf{max confidence}, which is the highest softmax probability of $\theta(x_i)$ across all classes; 
\textbf{entropy}, which is entropy of the output probabilities of $\theta(x_i)$; and 
\textbf{binary accuracy}, which indicates whether the model correctly predicts $y_i$ for $x_i$ (0 or 1).

Examples with high proxy values are learned earlier in the training process, tend to exhibit strong regularity within the overall data distribution, and contribute to better model generalization. 
Jiang et al. \cite{jiang2020characterizing} also demonstrated that the learning event proxies are highly correlated with the C-score, suggesting that these proxies effectively capture both the difficulty and regularity of examples during training.

\paragraph{Holdout retraining \cite{carlini2019distribution}} 
Proposed by Carlini et al. \cite{carlini2019distribution}, this proxy aims to capture the typicality or atypicality of data points.
% in a more efficient way than memorization scores. 
Given a model $\theta$ trained on $\dataset$ and an unseen test dataset $\testset$, the model $\theta$ is fine-tuned on $\testset$ to yield $\theta'$. 
For a data point $(x_i, y_i) \in \dataset$, the proxy computes the symmetric KL-divergence between the softmax probabilities of $\theta$ and $\theta'$.
Intuitively, a high proxy value indicates that the model's predictions for $(x_i, y_i)$ change significantly after fine-tuning, suggesting that $(x_i, y_i)$ is less typical of the overall data distribution.

\paragraph{Loss curvature \cite{garg2023memorization}}
This proxy was introduced by Garg et al. \cite{garg2023memorization} by using the curvature of the loss function around a given data point $(x_i, y_i)$ to approximate its memorization score. 
The curvature, as defined by Moosavi-Dezfooli et al. \cite{moosavi2019robustness}, is calculated from the derivative of the loss function with respect to the inputs, and the proxy value is obtained by averaging these curvatures over the course of training.
This proxy identifies data points where the model is more sensitive to perturbations, indicating stronger memorization.

\subsection{Approximate unlearning algorithms}\label{sec:rum}
% \textcolor{red}{Put all other algos in just one paragraph and then describe RUM as the SOTA, as per our NeurIPS submission - cite arxiv}.
\textbf{Fine-tune \cite{warnecke2021machine, golatkar2020eternal}} leverages "catastrophic forgetting" to reduce the model's knowledge of the forget set, achieving unlearning by continuing to train the original model $\theta_o$ on the retain set $\retainset$.
\textbf{NegGrad+ \cite{kurmanji2024towards}} extends the fine-tuning approach by applying gradient descent to the retain set $\retainset$, while treating the forget set $\forgetset$ differently using gradient ascent to encourage unlearning.
\textbf{L1-sparse \cite{liu2024model}} builds on Fine-tune by incorporating an L1 penalty to promote sparsity in the model weights.
\textbf{SalUn \cite{fan2023salun}} uses a random-label unlearning approach by assigning random labels to examples in $\forgetset$ and fine-tuning the model on both $\retainset$ and $\forgetset$ with these random labels \cite{golatkar2020eternal}. Specifically, SalUn identifies salient model weights and applies the random-label method only to those weights.

\textbf{RUM \cite{zhao2024makes}} is a meta-algorithm for unlearning proposed by Zhao et al. \cite{zhao2024makes}, which has empirically demonstrated significant improvements in unlearning performance across various existing algorithms. 
\begin{wrapfigure}{t}{0.5\textwidth}
  \begin{center}
    \includegraphics[scale=0.05]{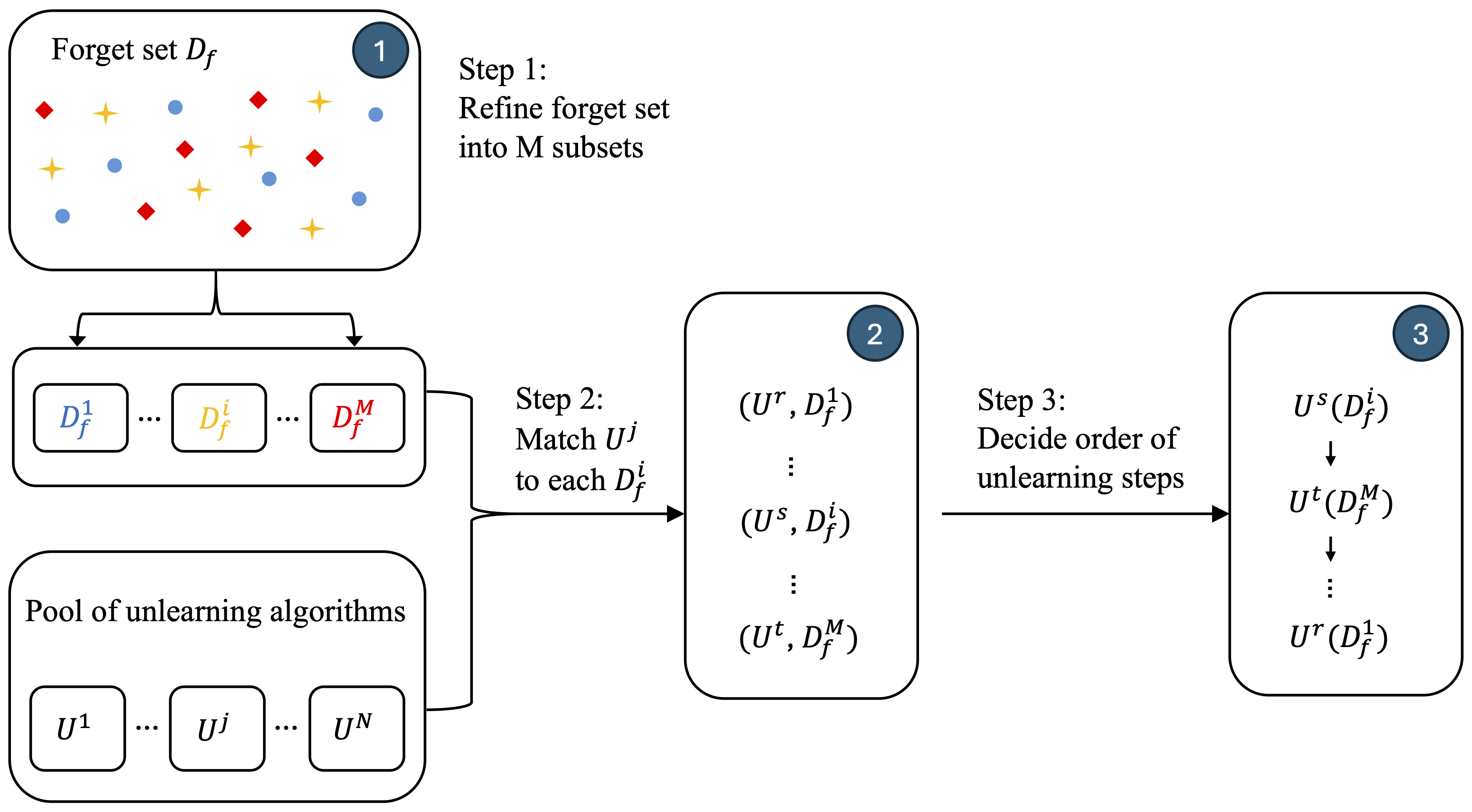}
  \end{center}
  \caption{Overview of RUM.}
   \label{fig:rum}
\end{wrapfigure}
An overview of RUM is shown in Figure \ref{fig:rum}.
RUM operates in two steps: (i) \emph{Refinement}, denoted by the function $\refine$, where the forget set $\forgetset$ is partitioned into $K$ homogeneous subsets based on the chosen inherent data property (e.g., memorization or its proxies): $\refine(\forgetset) = \{\forgetset^i\}_{i=1}^K$; (ii) \emph{Meta-Unlearning}, where an unlearning algorithm is selected from a pool of algorithms for each subset $\forgetset^i$, and then applied sequentially from the first subset to the last in a specified order.
Specifically, let $\mathcal{U}_1, \dots, \mathcal{U}_N$ represent a pool of state-of-the-art unlearning algorithms. For each subset $\forgetset^i$, we select an algorithm $\mathcal{U}^i \in {\mathcal{U}^1, \dots, \mathcal{U}^N}$ and perform $K$ unlearning steps in sequence. At step $i$, the selected algorithm $\mathcal{U}^i$ is applied to $\forgetset^i$, denoted as
$\mathcal{U}^i(\theta_o, \forgetset^i, \retainset^i) = \theta_u^i$, 
where $\theta_u^i$ is the model after unlearning step $i$, and $\retainset^i = \retainset \cup \{\forgetset^{i+1}, \dots, \forgetset^K\}$ is the retain set for step $i$, which includes $\retainset$ and all remaining subsets of $\forgetset$ yet to be unlearned. The process returns the final unlearned model $\theta_u^K$ after the last step.
In this study, we focus on RUM$^\mathcal{F}$ \cite{zhao2024makes}, where the refinement step is utilized by applying the same unlearning algorithm $\mathcal{U}$ sequentially to the subsets from $\refine(\forgetset)$, to investigate the impact of refinement alone.

\subsection{Evaluation metrics}\label{sec:eval}
% \subsection{ToW and ToW-MIA}
The unlearned model $\theta_u = \unlearn(\theta_o, \forgetset, \retainset)$ is expected to balance the forgetting of $\forgetset$ while preserving performance on $\retainset$ and generalizing well to the unseen test set $\testset$.
To assess this delicate balance between forgetting quality on $\forgetset$ and model utility performance on $\retainset$ and $\testset$, we adopt the "tug-of-war (\textbf{ToW})" metric, defined by Zhao et al. \cite{zhao2024makes} and introduce "\textbf{ToW-MIA}", a variant of ToW, to evaluate the unlearned model's performance from both accuracy and privacy perspectives.
The formal definitions of ToW and ToW-MIA are provided below:

\begin{equation*}
    \text{ToW}(\theta_u, \theta_r, \forgetset, \retainset, \testset) = (1 - \text{$\Delta$a}(\theta_u, \theta_r, \forgetset) ) \cdot 
    (1 - \text{$\Delta$a}(\theta_u, \theta_r, \retainset) ) \cdot 
    (1 - \text{$\Delta$a}(\theta_u, \theta_r, \testset) )
\end{equation*}

\begin{equation*}
    \text{ToW-MIA}(\theta_u, \theta_r, \forgetset, \retainset, \testset) = (1 - \text{$\Delta$m}(\theta_u, \theta_r, \forgetset) ) \cdot 
    (1 - \text{$\Delta$a}(\theta_u, \theta_r, \retainset) ) \cdot 
    (1 - \text{$\Delta$a}(\theta_u, \theta_r, \testset) )
% \label{eq:towmia}
\end{equation*}
where $\text{a}(\theta, \dataset) = \frac{1}{|\dataset|}{\sum_{(x,y) \in \dataset} [f(x; \theta) = y]}$ is the accuracy on $\dataset$ of a model $f$ parameterized by $\theta$ and $\text{$\Delta$a}(\theta_u, \theta_r, \dataset) = |\text{a}(\theta_u, \dataset) - \text{a}(\theta_r, \dataset)|$ is the absolute difference in accuracy between models $\theta_u$ and $\theta_r$ on $\dataset$. 
Similarly, $\text{m}(\theta, \dataset) = \frac{TN_{\forgetset}}{|\forgetset|}$ represents the MIA performance of a model with parameters $\theta$ on $\dataset$, and $\Delta \text{m}(\theta_u, \theta_r, \dataset) = |\text{m}(\theta_u, \dataset) - \text{m}(\theta_r, \dataset)|$ denote the absolute difference in MIA performance between models $\theta_u$ and $\theta_r$ on $\dataset$.
We used a commonly adopted MIA approach \citep{fan2023salun,liu2024model,zhao2024makes} for ToW-MIA, which involves training a binary classifier to distinguish between $\retainset$ and $\testset$ and then querying it with examples from $\forgetset$. See Section \ref{sec:mia_description} for further details on the MIA setup.

The only distinction between ToW and ToW-MIA lies in the "forgetting quality" term. ToW measures the relative accuracy difference on the forget set between the unlearned and retrained models, while ToW-MIA evaluates the relative difference in MIA performance on the forget set between the same models.
Both ToW and ToW-MIA reward unlearned models that closely match the performance of the retrained-from-scratch model. These metrics range from 0 to 1, with higher values indicating better unlearning performance.

\section{Profiles of memorization proxies}\label{sec:proxy-profile}
In this section, we evaluate the performance profiles of each proxy across two key dimensions: fidelity and efficiency.
% \paragraph{Fidelity} We assess the fidelity of each proxy 
\textbf{Fidelity} is assessed
by calculating the Spearman correlation coefficient between the proxy and memorization scores. The coefficient ranges from $[-1, 1]$, where a higher absolute value indicates a stronger correlation. 
% \paragraph{Efficiency} Efficiency 
\textbf{Efficiency} is measured by the extra computational time required to compute each proxy, in comparison to both computing memorization scores and retraining the original model $\theta_o$ from scratch (i.e., exact unlearning). 

\begin{table}[h]
\centering
\caption{
% \textcolor{red}{Use cifar100 and imagenet here and put cifar10 in appendix} 
Comparison of proxies based on Spearman correlation with memorization, computation time, and relative computation time percentages compared to memorization computing and retraining the model from scratch, evaluated on CIFAR-10 and CIFAR-100 datasets using ResNet-18 and ResNet-50 model architectures.}
\begin{subtable}{\textwidth}
\centering
\begin{tabularx}{0.9\textwidth}{l>{\centering\arraybackslash}X>{\centering\arraybackslash}X>{\centering\arraybackslash}X>{\centering\arraybackslash}X}
\toprule
Proxy & Spearman corr. (mem) & Computation time (s) & Comp. time \% (mem) & Comp. time \% (retrain) \\
\midrule
\textbf{Confidence} & -0.80 & 73.285 & 0.018\% & 17.123\% \\
Max confidence & -0.76 & 83.805 & 0.021\% & 19.581\% \\
Entropy & -0.75 & 115.838 & 0.029\% & 27.065\% \\
\textbf{Binary accuracy} & -0.71 & 72.154 & 0.018\% & 16.859\% \\
\textbf{Holdout retraining} & 0.67 & 69.263 & 0.017\% & 16.183\% \\
Loss curvature & 0.69 & 844.427 & 0.209\% & 197.298\% \\
\bottomrule
\end{tabularx}
\caption{CIFAR-10 with ResNet-18.}
% \label{tab:cifar10-resnet18}
\end{subtable}

\begin{subtable}{\textwidth}
\centering
\begin{tabularx}{0.9\textwidth}{l>{\centering\arraybackslash}X>{\centering\arraybackslash}X>{\centering\arraybackslash}X>{\centering\arraybackslash}X}
\toprule
Proxy & Spearman corr. (mem) & Computation time (s) & Comp. time \% (mem) & Comp. time \% (retrain) \\
\midrule
\textbf{Confidence} & -0.91 & 508.884 & 0.002\% & 8.175\% \\
Max confidence & -0.87 & 548.701 & 0.003\% & 8.815\% \\
Entropy & -0.80 & 734.146 & 0.004\% & 11.794\% \\
\textbf{Binary accuracy} & -0.89 & 441.257 & 0.002\% & 7.089\% \\
\textbf{Holdout retraining} & 0.62 & 209.236 & 0.001\% & 3.361\% \\
Loss curvature & 0.70 & 15142.780 & 0.074\% & 243.273\% \\
\bottomrule
\end{tabularx}
\caption{CIFAR-100 with ResNet-50.}
% \label{tab:cifar100-resnet50}
\end{subtable}
\label{tab:proxy-comparison}
\end{table}

The evaluation results for fidelity and efficiency are presented in Table \ref{tab:proxy-comparison}, with more detailed results, including distribution plots for the proxies versus memorization provided in the Section \ref{sec:proxy-distro}.
The results indicate that, among the learning event proxies, confidence and binary accuracy exhibit the highest Spearman correlation with memorization scores while also requiring the least computational time. Although holdout retraining exhibits only moderate correlation with memorization, it is far more efficient to compute and requires no intervention during model training compared to the other proxies.
Therefore, based on both fidelity and efficiency, we select the three best-performing proxies—\textbf{confidence}, \textbf{binary accuracy}, and \textbf{holdout retraining}—for further investigation into their impact on unlearning performance.

\section{How do proxies improve unlearning algorithms in RUM?}\label{sec:rum-experiment}
In this section, we explore the impact of integrating various proxies into RUM on existing unlearning algorithms. We assess the unlearning performance from both the \textbf{accuracy} and \textbf{privacy} perspectives, using ToW and ToW-MIA metrics, respectively. 

\paragraph{Experimental setup} \label{sec:setup}
We experiment with a refinement strategy based on proxy scores, setting $K = 3$ in RUM$^\mathcal{F}$ (see Section \ref{sec:rum}). The forget set $\forgetset$ consists of 3000 examples, divided into three subsets of $N = 1000$ examples each, representing the lowest, medium, and highest proxy values.
For each unlearning algorithm, we apply the refinement strategy \textbf{RUM$^\mathcal{F}$} following the same sequence as \cite{zhao2024makes}, unlearning in the order of low $\rightarrow$ medium $\rightarrow$ high memorization but using proxies in place of memorization scores. 
Additionally, we include two control setups from \cite{zhao2024makes}: \textbf{vanilla}, which unlearns the entire $\forgetset$ in one step, and \textbf{shuffle}, which uses random, equal-sized subsets of $\forgetset$ and operates sequentially on the three subsets.
We conduct the experiments on three dataset/architecture combinations: CIFAR-10 with ResNet-18, CIFAR-100 with ResNet-50, and Tiny-ImageNet with VGG-16. We evaluate the unlearning algorithm performance using ToW and ToW-MIA metrics. All the results are averaged over three runs with 95\% confidence intervals.

\begin{figure}[t]
\centering
\begin{subfigure}[b]{0.32\textwidth}
 \centering
\includegraphics[scale=0.25]{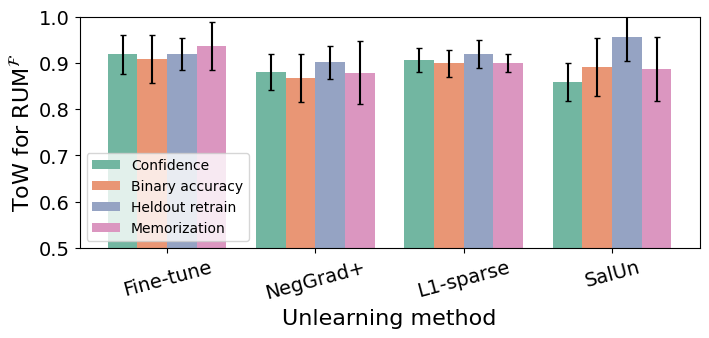}
 \caption{CIFAR-10 with ResNet-18}
 \label{fig:rum-tow-cifar10}
\end{subfigure}
%  \hfill
\begin{subfigure}[b]{0.32\textwidth}
\centering
\includegraphics[scale=0.25]{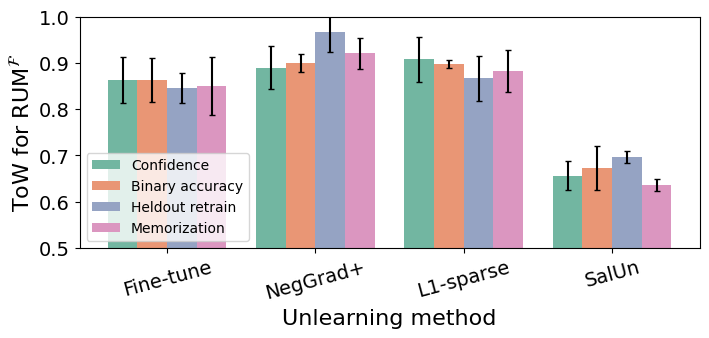}
\caption{CIFAR-100 with ResNet-50}
\label{fig:rum-tow-cifar100}
\end{subfigure}
\begin{subfigure}[b]{0.32\textwidth}
\centering
\includegraphics[scale=0.25]{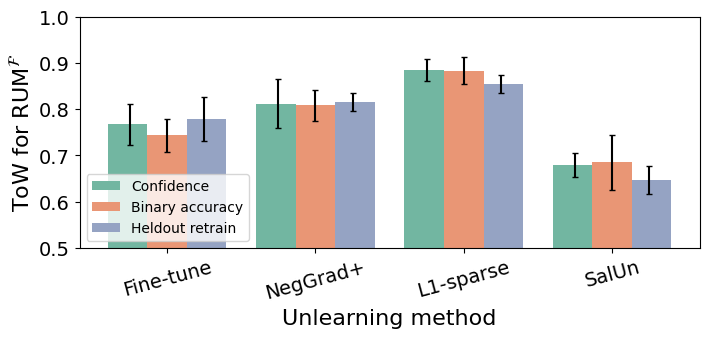}
\caption{Tiny-ImageNet with VGG-16}
\label{fig:rum-tow-tiny}
\end{subfigure}
% \caption{Uncovering the impact of three proxies (confidence, binary accuracy, holdout retraining) and memorization on unlearning performance in RUM$^\mathcal{F}$ from an accuracy perspective, evaluated using ToW (higher is better) across three datasets and model architectures.}
% \label{fig:rum-tow}
% \end{figure}

% \begin{figure}[t]
% \centering
\begin{subfigure}[b]{0.32\textwidth}
 \centering
\includegraphics[scale=0.25]{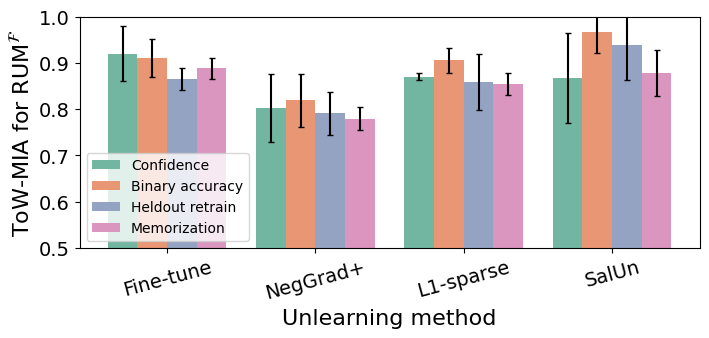}
 \caption{CIFAR-10 with ResNet-18}
 \label{fig:rum-towmia-cifar10}
\end{subfigure}
%  \hfill
\begin{subfigure}[b]{0.32\textwidth}
\centering
\includegraphics[scale=0.25]{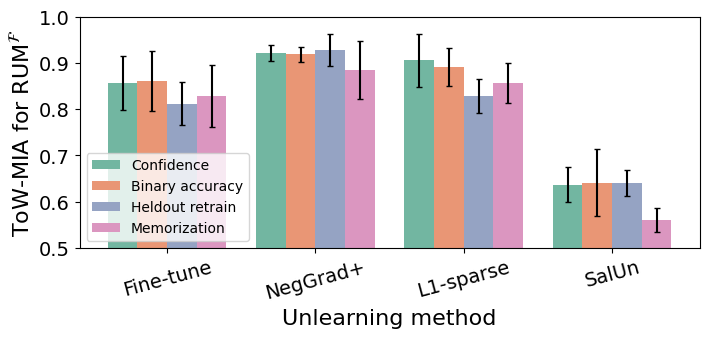}
\caption{CIFAR-100 with ResNet-50}
 \label{fig:rum-towmia-cifar100}
\end{subfigure}
\begin{subfigure}[b]{0.32\textwidth}
\centering
\includegraphics[scale=0.25]{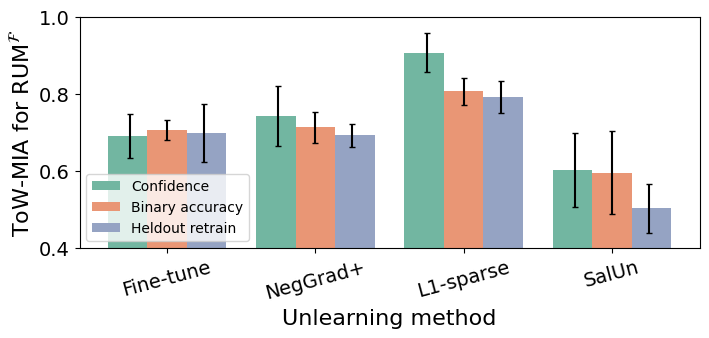}
\caption{Tiny-ImageNet with VGG-16}
 \label{fig:rum-towmia-tiny}
\end{subfigure}
\caption{Uncovering the impact of three proxies (confidence, binary accuracy, holdout retraining) and memorization on unlearning performance in RUM$^\mathcal{F}$, evaluated using ToW (Figures (a),(b),(c)) and ToW-MIA (Figures (d),(e),(f)) across three datasets and model architectures. Higher ToW/ToW-MIA values indicate better performance.}
\label{fig:rum-tow-towmia}
\end{figure}

\paragraph{Results and discussion}
The RUM$^\mathcal{F}$ results are illustrated in Figures \ref{fig:rum-tow-towmia}, with further details on the control experiments provided in the Section \ref{sec:rum-results}.
Moreover, Table \ref{tab:tow-towmia-rte} presents the ToW, ToW-MIA, and runtime results for each unlearning algorithm using different proxies in RUM$^\mathcal{F}$. Comprehensive results for all three datasets and architectures are available in Table \ref{tab:tow-towmia-rte-full}.

\begin{table}[t]
% \centering
% \begin{subtable}{\textwidth}
\centering
\caption{Comparison of unlearning algorithm performance using confidence, binary accuracy, and holdout retraining proxies, evaluated on CIFAR-100 with ResNet-50 (results for additional datasets/architectures are available in Table \ref{tab:tow-towmia-rte-full}).
Each algorithm $\mathcal{U}$ is applied in three different approaches: i) in one go ("vanilla"), ii) sequentially on a random partition of $\forgetset$ into three equal-sized subsets ("shuffle"), and iii) sequentially on three equal-sized subsets refined by $\refine$ ("RUM$^\mathcal{F}$").
Runtime indicates the time required for applying each algorithm $\mathcal{U}$ in the corresponding approach.}
\resizebox{\textwidth}{!}{
\begin{tabular}{lccc|ccc|ccc}
\toprule
\cmidrule(r){1-10}
& \multicolumn{3}{c}{Confidence} & \multicolumn{3}{c}{Binary accuracy} & \multicolumn{3}{c}{Holdout retraining} \\
{} & ToW ($\uparrow$) & ToW-MIA ($\uparrow$) & Runtime (s) & ToW ($\uparrow$) & ToW-MIA ($\uparrow$) & Runtime (s) & ToW ($\uparrow$) & ToW-MIA ($\uparrow$) & Runtime (s) \\
\midrule
\midrule
Retrain & 1.000 ± 0.000 & 1.000 ± 0.000 & {6254.604} & 1.000 ± 0.000 & 1.000 ± 0.000 & 6127.849 & 1.000 ± 0.000 & 1.000 ± 0.000 & 6430.247 \\
\midrule
Fine-tune RUM$^\mathcal{F}$ & 0.863 ± 0.049 & 0.857 ± 0.059 & 852.886 & 0.863 ± 0.048 & 0.861 ± 0.065 & 859.824 & 0.846 ± 0.032 & 0.812 ± 0.146 & 798.843 \\
Fine-tune shuffle & 0.674 ± 0.057 & 0.639 ± 0.079 & 843.585 & 0.671 ± 0.031 & 0.639 ± 0.062 & 865.150 & 0.714 ± 0.031 & 0.638 ± 0.045 & 852.442 \\
Fine-tune vanilla & 0.813 ± 0.061 & 0.880 ± 0.032 & 379.337 & 0.813 ± 0.015 & 0.868 ± 0.034 & 390.707 & 0.763 ± 0.028 & 0.803 ± 0.044 & 432.678 \\
\midrule
NegGrad+ RUM$^\mathcal{F}$ & 0.890 ± 0.047 & 0.922 ± 0.017 & 773.603 & 0.900 ± 0.020 & 0.919 ± 0.016 & 768.227 & 0.966 ± 0.042 & 0.928 ± 0.035 & 777.204 \\
NegGrad+ shuffle & 0.721 ± 0.020 & 0.712 ± 0.038 & 773.607 & 0.726 ± 0.007 & 0.719 ± 0.024 & 769.536 & 0.707 ± 0.016 & 0.618 ± 0.037 & 770.538 \\
NegGrad+ vanilla & 0.822 ± 0.025 & 0.836 ± 0.030 & 369.705 & 0.817 ± 0.053 & 0.821 ± 0.053 & 363.956 & 0.879 ± 0.046 & 0.790 ± 0.053 & 357.868 \\
\midrule
L1-sparse RUM$^\mathcal{F}$ & 0.908 ± 0.049 & 0.906 ± 0.057 & 783.477 & 0.897 ± 0.009 & 0.892 ± 0.041 & 782.627 & 0.867 ± 0.049 & 0.828 ± 0.037 & 769.910 \\
L1-sparse shuffle & 0.699 ± 0.031 & 0.670 ± 0.010 & 787.643 & 0.686 ± 0.016 & 0.658 ± 0.057 & 785.941 & 0.706 ± 0.005 & 0.613 ± 0.038 & 783.263 \\
L1-sparse vanilla & 0.796 ± 0.099 & 0.797 ± 0.084 & 395.368 & 0.771 ± 0.112 & 0.795 ± 0.094 & 396.259 & 0.770 ± 0.024 & 0.730 ± 0.115 & 397.543 \\
\midrule
SalUn RUM$^\mathcal{F}$ & 0.656 ± 0.031 & 0.636 ± 0.038 & 791.166 & 0.673 ± 0.048 & 0.641 ± 0.072 & 793.327 & 0.696 ± 0.013 & 0.640 ± 0.129 & 793.793 \\
SalUn shuffle & 0.603 ± 0.052 & 0.541 ± 0.055 & 792.552 & 0.636 ± 0.030 & 0.591 ± 0.035 & 795.967 & 0.581 ± 0.039 & 0.488 ± 0.045 & 793.672 \\
SalUn vanilla & 0.633 ± 0.043 & 0.543 ± 0.186 & 417.232 & 0.651 ± 0.050 & 0.705 ± 0.035 & 421.418 & 0.617 ± 0.030 & 0.478 ± 0.163 & 396.784 \\
\bottomrule
\end{tabular}}
% \caption{CIFAR-100 with ResNet-50}
% \end{subtable}
\label{tab:tow-towmia-rte}
\end{table}

% \paragraph{Discussion}
The experimental results show that all proxies can improve the performance of unlearning algorithms in terms of both accuracy and privacy when integrated into RUM$^\mathcal{F}$, with some proxies even outperforming memorization. 
This outcome is expected, as memorization and proxies capture similar but nuanced aspects of the data. 
Memorization reflects the model's behavior when trained with or without a specific example, while proxies like learning events measure how easily the model learns an example during training. The holdout retraining proxy, in contrast, assesses whether an example is well-represented by others, particularly in identifying atypical data points.

From Figure \ref{fig:rum-tow-towmia}, we observe that SalUn underperforms on CIFAR-100 and Tiny-ImageNet compared to other unlearning algorithms. 
This can be attributed to the use of data augmentation on these datasets (but not on CIFAR-10), which makes these models more robust to noise. As SalUn is a relabelling-based algorithm that introduces noisy labels to facilitate unlearning, its effectiveness is reduced in models that are more resilient to noise.
As a result, SalUn achieves incomplete unlearning, leaving the influence of the forget set partially intact and making the model more vulnerable to MIA on the forget set data, compared to other unlearning algorithms.

Another notable observation from Figure \ref{fig:rum-tow-cifar10} is that holdout retraining outperforms other proxies on CIFAR-10, even surpassing memorization in most cases. 
This is likely because holdout retraining is the only proxy that has access to all data examples (both the training and test sets) during computation, whereas the other proxies only rely on the training set. This broader exposure may give the model a better grasp of an example's atypicality by leveraging a larger image pool. 
However, when data augmentation is applied (Figures \ref{fig:rum-tow-cifar100} and \ref{fig:rum-tow-tiny}), synthetic variations of the training images are generated, effectively increasing the number of training examples, reducing the relative advantage of holdout retraining.
% This reduces the relative advantage of holdout retraining, as the expanded training set diminishes the importance of any individual data point. 
This suggests that holdout retraining is highly effective when no data augmentation is used and could be a strong option for RUM in such cases.
Conversely, Figures \ref{fig:rum-towmia-cifar10}, \ref{fig:rum-towmia-cifar100} and \ref{fig:rum-towmia-tiny} shows that no single proxy consistently outperforms the others across all scenarios. This indicates that the proxies may have similar effects on privacy when integrated into RUM$^\mathcal{F}$, with no one proxy offering a universal advantage.

In terms of efficiency, Table \ref{tab:tow-towmia-rte} and Table \ref{tab:tow-towmia-rte-full} show that while RUM$^\mathcal{F}$ requires approximately twice the runtime of the vanilla approach, it still takes significantly less time than retraining from scratch. 
This efficiency advantage becomes even more evident with larger datasets. Given the substantial performance gains RUM$^\mathcal{F}$ offers with reasonable extra overhead, it emerges as a highly promising unlearning approach.
Regarding unlearning algorithms, we observe that Fine-tune RUM$^\mathcal{F}$ requires more runtime on larger datasets such as CIFAR-100 and Tiny-ImageNet compared to other baselines. However, on smaller datasets like CIFAR-10, NegGrad+ RUM$^\mathcal{F}$ takes more runtime than the other methods.
When comparing proxies, no single proxy stands out as significantly more efficient when applying an algorithm $\mathcal{U}$ across different approaches. However, as discussed in Section \ref{sec:proxy-profile}, holdout retraining is more computationally efficient than other proxies and does not require any intervation during the training process, making it a strong candidate as a proxy for memorization.

\paragraph{Stability analysis} \label{sec:stability}
One may reasonably expect that memorization scores may change after successive unlearning operations, analogously to the "onion effect" \cite{carlini2022privacy}.
This raises issues of stability of the improvements achieved by memorization proxies in sequential unlearning.
We shed light on this issue by examining changes in unlearning performance before and after each unlearning step across multiple unlearning iterations, in order to understand the cumulative effects of unlearning over time and the impact of memorization proxies on this. We apply multiple sequential unlearning steps and track performance in terms of accuracy and privacy, which are evaluated using ToW and ToW-MIA, respectively.
% \textcolor{red}{Furthermore, A stable proxy should exhibit minimal variation in its values across the unlearning process, ensuring consistency in its representation of the data’s influence on the model. This is particularly important for maintaining reliable performance when proxies are used to guide the unlearning process.}
% Analyzing this progression can provide insights into the potential cumulative effects of unlearning and help identify whether the process leads to improved model behavior or introduces unintended trade-offs.
We use NegGrad+ as the unlearning algorithm and experiment on CIFAR-10/ResNet-18 and Tiny-ImageNet/VGG-16, as described in Section \ref{sec:setup} for both RUM$^\mathcal{F}$ and vanilla approaches.
%We apply unlearning sequentially over 5 steps and compare the performance of RUM$^\mathcal{F}$ and vanilla across these iterations.
After each unlearning step $n$, we recalculate the proxy values and reapply the partitioning procedure based on the updated proxy values, which involves selecting three subsets (lowest, medium, and highest proxy values) of 1000 examples each, which form the forget set of size 3,000 for step $n+1$.

\begin{figure}[t]
\centering
\begin{subfigure}[b]{0.49\textwidth}
 \centering
\includegraphics[scale=0.3]{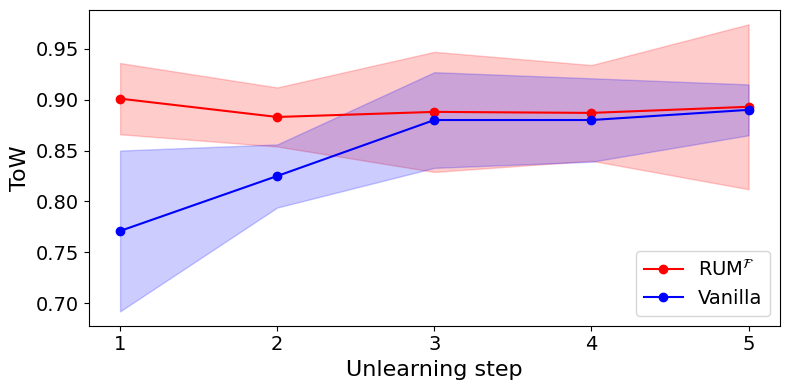}
 \caption{CIFAR-10 with ResNet-18}
 % \label{fig:rum-tow-cifar10}
\end{subfigure}
%  \hfill
\begin{subfigure}[b]{0.49\textwidth}
\centering
\includegraphics[scale=0.3]{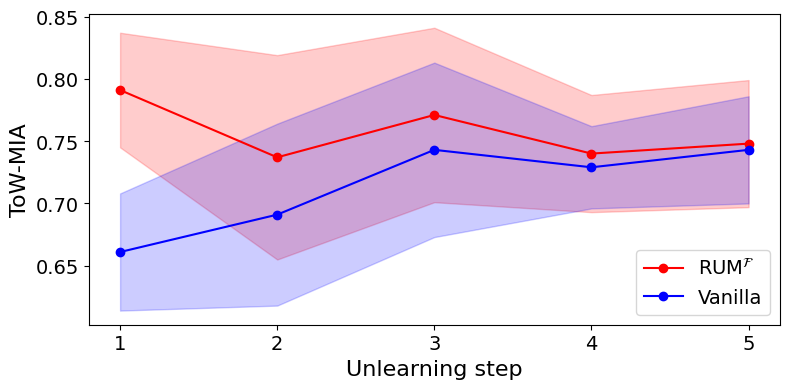}
\caption{CIFAR-10 with ResNet-18}
% \label{fig:rum-tow-cifar100}
\end{subfigure}

\begin{subfigure}[b]{0.49\textwidth}
 \centering
\includegraphics[scale=0.3]{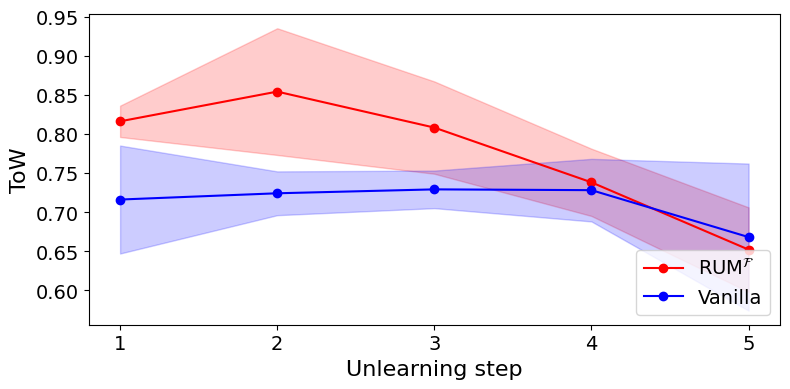}
 \caption{Tiny-ImageNet with VGG-16}
 % \label{fig:rum-towmia-cifar10}
\end{subfigure}
%  \hfill
\begin{subfigure}[b]{0.49\textwidth}
\centering
\includegraphics[scale=0.3]{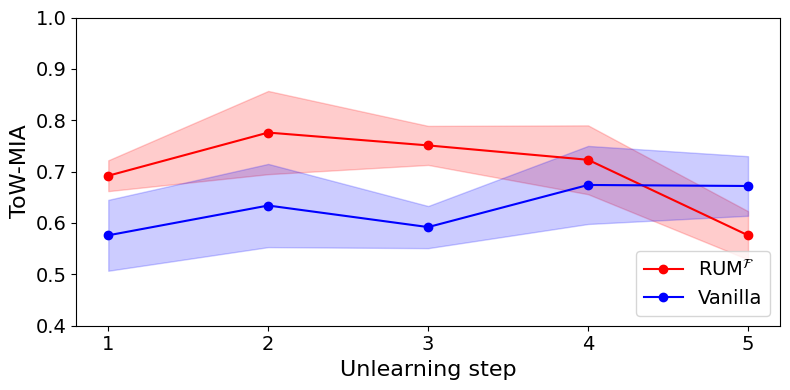}
\caption{Tiny-ImageNet with VGG-16}
 % \label{fig:rum-towmia-cifar100}
\end{subfigure}
\caption{Cumulative performance changes over 5-step sequential unlearning in RUM$^\mathcal{F}$ and vanilla using NegGrad+ as the baseline, evaluated by ToW (Figures (a), (c)) and ToW-MIA (Figures (b), (d)) across two datasets and model architectures. Higher ToW/ToW-MIA values indicate better performance.}
\label{fig:seq-tow-towmia}
\end{figure}

Figure \ref{fig:seq-tow-towmia} and Table \ref{tab:seq-tow-towmia} show the results. 
For CIFAR-10 (Figures (a) and (b)), RUM$^\mathcal{F}$ remains relatively stable across both ToW and ToW-MIA metrics. Vanilla shows an upward trend of improved performance. The performance gap between them narrows with each subsequent unlearning step.
The upward trend in vanilla may result from the sequential removal of highly memorized examples, which are more difficult to unlearn according to \cite{zhao2024makes}. As these challenging examples are gradually unlearned, the unlearning problem becomes easier, leading to improved performance of the vanilla version.
Results for Tiny-ImageNet/VGG-16 are shown in Figures (c) and (d). 
We see for both that RUM$^\mathcal{F}$ performance is declining with successive iterations, whereas vanilla tends to either improve or be less affected.
The main reason for this is that the superiority of RUM$^\mathcal{F}$ lies in its ability to distinguish high- versus low-memorized data examples. Successive unlearning steps have as a result the removal of most highly-memorized examples, hence reducing the possible improvement gains of RUM$^\mathcal{F}$, as forget sets become after a given point inherently homogenized.
Please note, however, that in typical application settings, only a very small percentage of the dataset will be unlearned (unlike our experimental setup where we ended up unlearning a large percentage of the dataset) which will leave the memorization distribution unaffected.
%\textcolor{red}{BUT what does this say about the stability of proxies???} \textcolor{blue}{Can we say that: Therefore, we see that the use of proxies does not affect adversely the performance of RUM? Can we have numbers for what RUM performance would be for the last unlearning step if we had the exact memorization scores? ie compute exact mem scores after step 4 and apply RUM-F using those for step 5 and compare?}

To confirm that the drop in RUM$^\mathcal{F}$ performance is due to the removal of most outlier examples with very high memorization, we calculated the Gini coefficient \cite{bendel1989comparison}, along with Lorenz curves \cite{fellman1976effect}. Combined they offer an explainable metric for the skewness (inequity) of how a variable $X$ (say total income or total memorization scores) is distributed across a population $P$ (say people or data examples).  A point ($\alpha, \beta$) on a Lorenz curve depicts that $\beta \%$ of $X$ comes from $\alpha \%$ of $P$. The Gini coefficient measures how far the curve is from a perfectly equal distribution (where any $\alpha \%$ of $X$ contributes $\alpha \%$ to $P$) and 
how close it is to a totally skewed distribution (where a single member of the population accounts for 100\% of the value of $X$).
Gini values range from 0 to 1, where higher values indicate greater distribution skewness (inequality). 
Gini and Lorenz curves are appropriate in our setting which is dominated by a few examples contributing more to the cumulative memorization score of all examples. We hypothesize that this distribution changes over time, after successive unlearning steps. To track these changes, we calculate the Gini values before and after each unlearning step (see Table \ref{tab:gini}). The results show a marked decrease in the Gini coefficient for both vanilla and RUM$^\mathcal{F}$, indicating a progressively less skewed distribution. This reduction in skewness reflects the removal of highly memorized, long-tailed examples, diminishing the advantage of RUM$^\mathcal{F}$ over steps. 
Figure \ref{fig:proxy-distro-seq} further visualizes the proxy value distribution across unlearning steps.

%This may be due to Tiny-ImageNet’s larger size, making it less sensitive to changes from sequentially unlearning forget sets.  Additionally, we observe a performance decline over the unlearning steps on Tiny-ImageNet, which could be attributed to catastrophic forgetting \cite{french1999catastrophic}. This decline is more pronounced in RUM$^\mathcal{F}$, where multiple sequential steps within each unlearning iteration exacerbate this side-effect. Consequently, the "catastrophic forgetting" outweighs the benefits of the homogenization of the forget set, as highlighted in \cite{zhao2024makes}.

\begin{table}[t]
\centering
\caption{Gini coefficient of proxy values over 5 sequential unlearning steps. A higher Gini indicates greater skewness in the distribution. Step 0 represents the state before any unlearning is applied.}
\centering
\resizebox{0.8\textwidth}{!}{
\begin{tabular}{lcc|cc}
\toprule
\cmidrule(r){1-5}
& \multicolumn{2}{c}{CIFAR-10 / ResNet-18} & \multicolumn{2}{c}{Tiny-ImageNet / VGG-16} \\
{} & NegGrad+ RUM$^\mathcal{F}$ & NegGrad+ vanilla & NegGrad+ RUM$^\mathcal{F}$ & NegGrad+ vanilla \\
\midrule
Step 0 & 0.788 & 0.788 & 0.629 & 0.629 \\
Step 1 & 0.742 & 0.758 & 0.547 & 0.634 \\
Step 2 & 0.683 & 0.751 & 0.487 & 0.616 \\
Step 3 & 0.697 & 0.730 & 0.427 & 0.596 \\
Step 4 & 0.676 & 0.726 & 0.386 & 0.440 \\
Step 5 & 0.642 & 0.723 & 0.365 & 0.403 \\
\bottomrule
\end{tabular}}
\label{tab:gini}
\end{table}

\section{Conclusion}
A new class of memorization-based algorithms has emerged that significantly improves unlearning quality while preserving model utility. However, their heavy reliance on exact memorization scores à la Feldman, which are notoriously computationally expensive, limits their scalability. This paper conducts an in-depth analysis of the performance of such unlearning algorithms when using memorization proxies instead. Our findings demonstrate that these substantial performance gains can indeed be achieved efficiently and at scale. 
Specifically, integrating memorization proxies into RUM$^\mathcal{F}$ enhances unlearning performance from both accuracy and privacy perspectives, with up to a 30\% improvement in accuracy and up to a 46\% improvement in privacy compared to baseline methods. 
% ; (ii) data augmentation negatively impacts relabeling-based unlearning algorithms like SalUn, while avoiding data augmentation during training can give the holdout retraining proxy an advantage over others \textcolor{red}{Not sure this is important enough to mention in conclusions}; 
Among the proxies evaluated, although no single proxy consistently outperforms others across all scenarios, holdout retraining stands out for its efficiency—requiring up to 99.98\% less runtime than computing exact memorization scores—and its practicality, as it requires no intervention during model training. This significant reduction in computational cost makes proxies a scalable alternative for memorization-based unlearning, achieving near-equivalent performance without the prohibitive overhead of exact memorization scores.
Furthermore, while successive unlearning steps change the underlying memorization score distributions, the performance gains offered by RUM$^\mathcal{F}$ using memorization proxies appear to hold, up to a point where memorization score distribution becomes less skewed for RUM to have any performance impact. The main conclusion therefore is that memorization-based unlearning algorithms can now offer scalability and efficiency along with their great unlearning accuracy and privacy performance. 
These findings not only highlight the scalability of memorization-based unlearning but also pave the way for new possibilities in efficient machine unlearning.
% With respect to comparing proxies, although no single proxy consistently outperforms others across all scenarios, holdout retraining emerges as a strong candidate due to its high computational efficiency \textcolor{red}{quantify this} and lack of required intervention during the training process \textcolor{red}{quantify this}.
% We believe these findings make memorization-based unlearning scalable and practical, \textcolor{red}{explain why - quantify how much more expensive it is without proxies and what we gain using proxies explicitly - this IS the main thing!}. 
% These findings we hope open new possibilities for machine unlearning.

\begin{ack}
We would like to express our gratitude to Eleni Triantafillou, Meghdad Kurmanji, and George-Octavian Barbulescu for their valuable feedback and contributions throughout this work.
% Use unnumbered first level headings for the acknowledgments. All acknowledgments
% go at the end of the paper before the list of references. Moreover, you are required to declare
% funding (financial activities supporting the submitted work) and competing interests (related financial activities outside the submitted work).
% More information about this disclosure can be found at: \url{https://neurips.cc/Conferences/2024/PaperInformation/FundingDisclosure}.

% Do {\bf not} include this section in the anonymized submission, only in the final paper. You can use the \texttt{ack} environment provided in the style file to autmoatically hide this section in the anonymized submission.
\end{ack}

\bibliographystyle{plain}
\bibliography{references}

\begin{thebibliography}{10}

\bibitem{arpit2017closer}
Devansh Arpit, Stanis{\l}aw Jastrz{\k{e}}bski, Nicolas Ballas, David Krueger, Emmanuel Bengio, Maxinder~S Kanwal, Tegan Maharaj, Asja Fischer, Aaron Courville, Yoshua Bengio, et~al.
\newblock A closer look at memorization in deep networks.
\newblock In {\em International conference on machine learning}, pages 233--242. PMLR, 2017.

\bibitem{barbulescu2024each}
George-Octavian Barbulescu and Peter Triantafillou.
\newblock To each (textual sequence) its own: Improving memorized-data unlearning in large language models.
\newblock In {\em International conference on machine learning}, 2024.

\bibitem{bendel1989comparison}
RB~Bendel, SS~Higgins, JE~Teberg, and DA~Pyke.
\newblock Comparison of skewness coefficient, coefficient of variation, and gini coefficient as inequality measures within populations.
\newblock {\em Oecologia}, 78:394--400, 1989.

\bibitem{cao2015towards}
Yinzhi Cao and Junfeng Yang.
\newblock Towards making systems forget with machine unlearning.
\newblock In {\em 2015 IEEE symposium on security and privacy}, pages 463--480. IEEE, 2015.

\bibitem{carlini2019distribution}
Nicholas Carlini, Ulfar Erlingsson, and Nicolas Papernot.
\newblock Distribution density, tails, and outliers in machine learning: Metrics and applications.
\newblock {\em arXiv preprint arXiv:1910.13427}, 2019.

\bibitem{carlini2022privacy}
Nicholas Carlini, Matthew Jagielski, Chiyuan Zhang, Nicolas Papernot, Andreas Terzis, and Florian Tramer.
\newblock The privacy onion effect: Memorization is relative.
\newblock {\em Advances in Neural Information Processing Systems}, 35:13263--13276, 2022.

\bibitem{fan2023salun}
Chongyu Fan, Jiancheng Liu, Yihua Zhang, Dennis Wei, Eric Wong, and Sijia Liu.
\newblock Salun: Empowering machine unlearning via gradient-based weight saliency in both image classification and generation.
\newblock {\em arXiv preprint arXiv:2310.12508}, 2023.

\bibitem{feldman2020does}
Vitaly Feldman.
\newblock Does learning require memorization? a short tale about a long tail.
\newblock In {\em Proceedings of the 52nd Annual ACM SIGACT Symposium on Theory of Computing}, pages 954--959, 2020.

\bibitem{feldman2020neural}
Vitaly Feldman and Chiyuan Zhang.
\newblock What neural networks memorize and why: Discovering the long tail via influence estimation.
\newblock {\em Advances in Neural Information Processing Systems}, 33:2881--2891, 2020.

\bibitem{fellman1976effect}
Johan Fellman.
\newblock The effect of transformations on lorenz curves.
\newblock {\em Econometrica (pre-1986)}, 44(4):823, 1976.

\bibitem{garg2023memorization}
Isha Garg, Deepak Ravikumar, and Kaushik Roy.
\newblock Memorization through the lens of curvature of loss function around samples.
\newblock {\em arXiv preprint arXiv:2307.05831}, 2023.

\bibitem{golatkar2020eternal}
Aditya Golatkar, Alessandro Achille, and Stefano Soatto.
\newblock Eternal sunshine of the spotless net: Selective forgetting in deep networks.
\newblock In {\em Proceedings of the IEEE/CVF Conference on Computer Vision and Pattern Recognition}, pages 9304--9312, 2020.

\bibitem{hoofnagle2019european}
Chris~Jay Hoofnagle, Bart Van Der~Sloot, and Frederik~Zuiderveen Borgesius.
\newblock The european union general data protection regulation: what it is and what it means.
\newblock {\em Information \& Communications Technology Law}, 28(1):65--98, 2019.

\bibitem{jiang2020characterizing}
Ziheng Jiang, Chiyuan Zhang, Kunal Talwar, and Michael~C Mozer.
\newblock Characterizing structural regularities of labeled data in overparameterized models.
\newblock {\em arXiv preprint arXiv:2002.03206}, 2020.

\bibitem{kurmanji2024towards}
Meghdad Kurmanji, Peter Triantafillou, Jamie Hayes, and Eleni Triantafillou.
\newblock Towards unbounded machine unlearning.
\newblock {\em Advances in Neural Information Processing Systems}, 36, 2023.

\bibitem{liu2024model}
Jiancheng Liu, Parikshit Ram, Yuguang Yao, Gaowen Liu, Yang Liu, PRANAY SHARMA, Sijia Liu, et~al.
\newblock Model sparsity can simplify machine unlearning.
\newblock {\em Advances in Neural Information Processing Systems}, 36, 2024.

\bibitem{moosavi2019robustness}
Seyed-Mohsen Moosavi-Dezfooli, Alhussein Fawzi, Jonathan Uesato, and Pascal Frossard.
\newblock Robustness via curvature regularization, and vice versa.
\newblock In {\em Proceedings of the IEEE/CVF Conference on Computer Vision and Pattern Recognition}, pages 9078--9086, 2019.

\bibitem{nguyen2022survey}
Thanh~Tam Nguyen, Thanh~Trung Huynh, Phi~Le Nguyen, Alan Wee-Chung Liew, Hongzhi Yin, and Quoc Viet~Hung Nguyen.
\newblock A survey of machine unlearning.
\newblock {\em arXiv preprint arXiv:2209.02299}, 2022.

\bibitem{rosen2011right}
Jeffrey Rosen.
\newblock The right to be forgotten.
\newblock {\em Stan. L. Rev. Online}, 64:88, 2011.

\bibitem{toneva2018empirical}
Mariya Toneva, Alessandro Sordoni, Remi Tachet~des Combes, Adam Trischler, Yoshua Bengio, and Geoffrey~J Gordon.
\newblock An empirical study of example forgetting during deep neural network learning.
\newblock {\em arXiv preprint arXiv:1812.05159}, 2018.

\bibitem{triantafillou2024we}
Eleni Triantafillou, Peter Kairouz, Fabian Pedregosa, Jamie Hayes, Meghdad Kurmanji, Kairan Zhao, Vincent Dumoulin, Julio~Jacques Junior, Ioannis Mitliagkas, Jun Wan, Lisheng Sun~Hosoya, Sergio Escalera, Gintare~Karolina Dziugaite, Peter Triantafillou, and Isabelle Guyon.
\newblock Are we making progress in unlearning? findings from the first neurips unlearning competition.
\newblock {\em arXiv preprint arXiv:2406.09073}, 2024.

\bibitem{warnecke2021machine}
Alexander Warnecke, Lukas Pirch, Christian Wressnegger, and Konrad Rieck.
\newblock Machine unlearning of features and labels.
\newblock {\em arXiv preprint arXiv:2108.11577}, 2021.

\bibitem{zhao2024makes}
Kairan Zhao, Meghdad Kurmanji, George-Octavian B{\u{a}}rbulescu, Eleni Triantafillou, and Peter Triantafillou.
\newblock What makes unlearning hard and what to do about it.
\newblock {\em arXiv preprint arXiv:2406.01257 (to appear in NeurIPS 2024)}, 2024.

\end{thebibliography}

%%%%%%%%%%%%%%%%%%%%%%%%%%%%%%%%%%%%%%%%%%%%%%%%%%%%%%%%%%%%

\clearpage

\appendix

\section{Appendix / supplemental material}
% \section*{Supplementary Material}

\subsection{Implementation details} \label{sec:implementation_details}

We use three settings with different datasets and model architectures for our evaluation: CIFAR-10 with ResNet-18, CIFAR-100 with ResNet-50, and Tiny-ImageNet with VGG-16.
All experiments were implemented in PyTorch, and conducted on Nvidia RTX A5000 GPUs.
Specific details for training the original models can be found in Table \ref{tab:training-details}.

\begin{table}[h]
\centering
\begin{threeparttable} % Begin the threeparttable environment
\caption{Training configurations across three settings.}
\begin{tabularx}{\textwidth}{l>{\centering\arraybackslash}X>{\centering\arraybackslash}X>{\centering\arraybackslash}X}
\toprule
Dataset & CIFAR-10 & CIFAR-100 & Tiny-ImageNet \\
\midrule
Number of classes & 10 & 100 & 200 \\
Training set size & 45,000 & 45,000 & 100,000 \\
Architecture & ResNet-18 & ResNet-50 & VGG-16 \\
Optimizer & SGD & SGD & SGD \\
Base learning rate & 0.1 & 0.1 & 0.1 \\
Learning rate scheduler & CosineAnnealingLR & MultiStepLR\tnote{*} & CosineAnnealingLR \\
Batch size & 256 & 256 & 256 \\
Epochs & 30 & 150 & 100 \\
Momentum & 0.9 & 0.9 & 0.9 \\
Weight decay & $5\times10^{-4}$ & $5\times10^{-4}$ & $5\times10^{-4}$ \\
Data augmentation & None & Random Crop + Horizontal Flip & Random Crop + Horizontal Flip \\
\bottomrule
\end{tabularx}
\begin{tablenotes}
\footnotesize
\item[*] The learning rate was initialized at 0.1 and decayed by a factor of 0.2 at 60 and 120 epochs.
\end{tablenotes}
\label{tab:training-details}
\end{threeparttable} % End the threeparttable environment
\end{table}

\paragraph{Training details for machine unlearning}
% retrain: same training procedure as original model but without forget set, on retain set only
% fine-tune: 5-10 epochs, lr[0.001,0.1]
% l1-sparse: 5-10 epochs, lr[0.001,0.1], $\gamma=[10^{-5},5*10^{-4}]$
% NegGrad+: $\beta$[0.9,0.99], 5 epochs, lr[0.001,0.05]
% SalUn: 5-10 epochs, [0.005,0.1], sparsity ratios in [0.3,0.7]
In the unlearning process, several state-of-the-art algorithms were employed, each with carefully tuned hyperparameters to ensure optimal performance across different datasets and architectures. 
Retrain-from-scratch follows the same training procedure as the original model but is performed solely on the retain set $\retainset$, excluding the forget set $\forgetset$. 
Fine-tune involves training the model for 5 to 10 epochs with a learning rate ranging from 0.001 to 0.1. 
L1-Sparse also runs for 5 to 10 epochs with a learning rate between 0.001 and 0.1, using a sparsity regularization parameter $\gamma$ in the range of $10^{-5}$ to $5 \times 10^{-4}$. 
NegGrad+ is executed for 5 epochs, using a learning rate between 0.001 and 0.05 and a $\beta$ parameter ranging from 0.9 to 0.99. 
SalUn operates for 5 to 10 epochs with a learning rate between 0.005 and 0.1 and applies sparsity ratios between 0.3 and 0.7. 
These varied hyperparameters allow each algorithm to efficiently facilitate the unlearning process under different conditions.

\subsection{Description of MIA} \label{sec:mia_description}
In this study, we adopted a commonly used MIA from prior work \cite{liu2024model, fan2023salun, zhao2024makes} to evaluate unlearning performance from a privacy perspective.
To measure MIA performance, we first sample equal-sized data from the retain set $\retainset$ and test set $\testset$ to train a binary classifier that distinguishes between data points involved in training and those that were not. After applying an unlearning algorithm, we apply this classifier to the unlearned model $\theta_u$ on the forget set $\forgetset$. If an example has been effectively "forgotten", the classifier should identify it as "non-training" data, as if it came from $\testset$.

We define "training" data as the positive class and "non-training" data as the negative class. The MIA score is calculated as the proportion of true negatives—$\forgetset$ examples correctly classified as "non-training." A score closer to 1 indicates more effective unlearning. Ideally, the MIA score should match that of retraining-from-scratch, but due to the similarity between the forget set and retain set, some examples may still be classified as "training." To account for this, we calculate the "MIA gap", the absolute difference between the MIA score of the unlearned model and that of retraining-from-scratch, and incorporate it into the "ToW-MIA" evaluation (Section \ref{sec:eval}) as a measure of "forgetting quality", where a smaller MIA gap indicates better unlearning performance.

% as shown in Equation \ref{defn:MIA}
% \begin{equation}
% \text{MIA Score} = \frac{TN_{\forgetset}}{|\forgetset|},
% \label{defn:MIA}
% \end{equation}

\subsection{Detailed results} \label{sec:results}
\subsubsection{Distribution of memorization vs. proxies} 
\label{sec:proxy-distro}
% \paragraph{Distribution of memorization vs. proxies}

\begin{figure}[t]
\centering
% Group 1: CIFAR-10
\begin{minipage}[b]{\textwidth}
\centering
\captionsetup{type=figure} % Allows for a group caption
\begin{subfigure}[b]{0.32\textwidth}
 \centering
\includegraphics[scale=0.2]{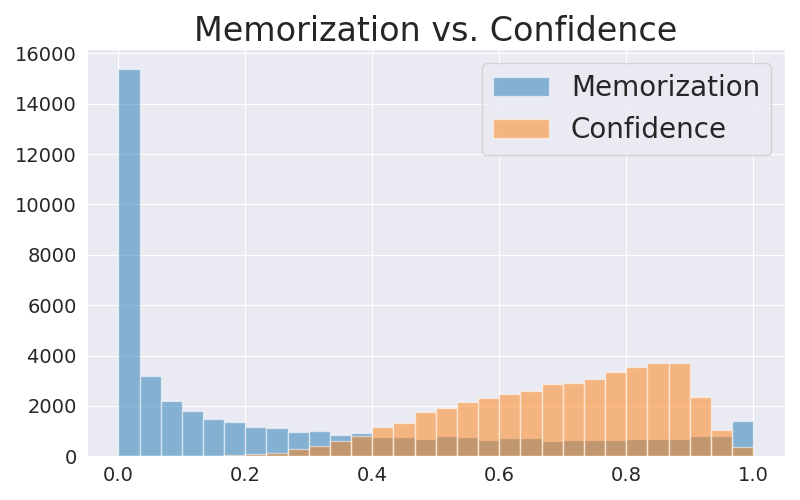}
 % \caption{Memorization vs Confidence}
\end{subfigure}
\begin{subfigure}[b]{0.32\textwidth}
\centering
\includegraphics[scale=0.2]{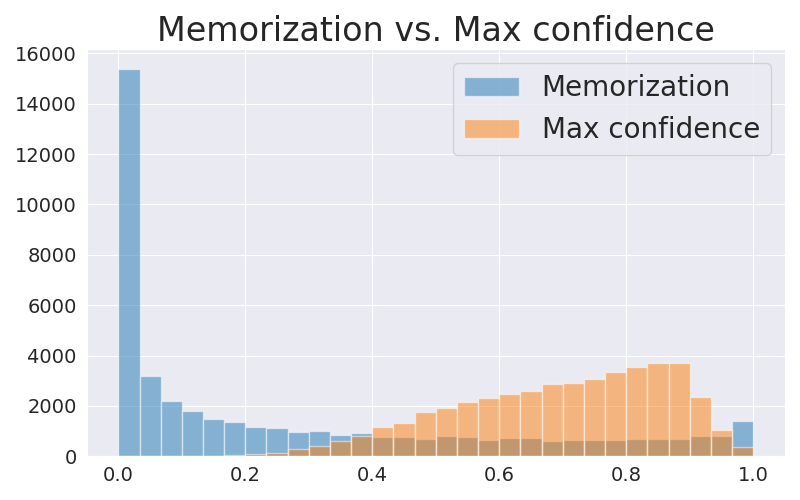}
 % \caption{Memorization vs Holdout retraining}
\end{subfigure}
\begin{subfigure}[b]{0.32\textwidth}
\centering
\includegraphics[scale=0.2]{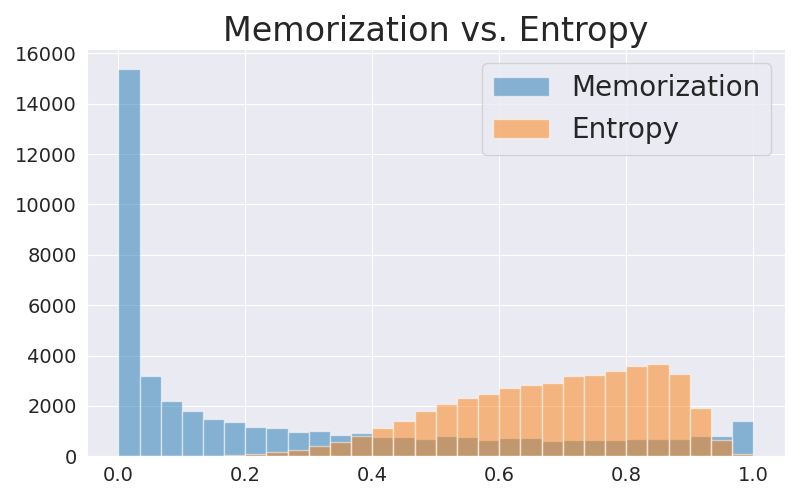}
% \caption{Tiny-ImageNet with VGG-16}
\end{subfigure}
\begin{subfigure}[b]{0.32\textwidth}
\centering
\includegraphics[scale=0.2]{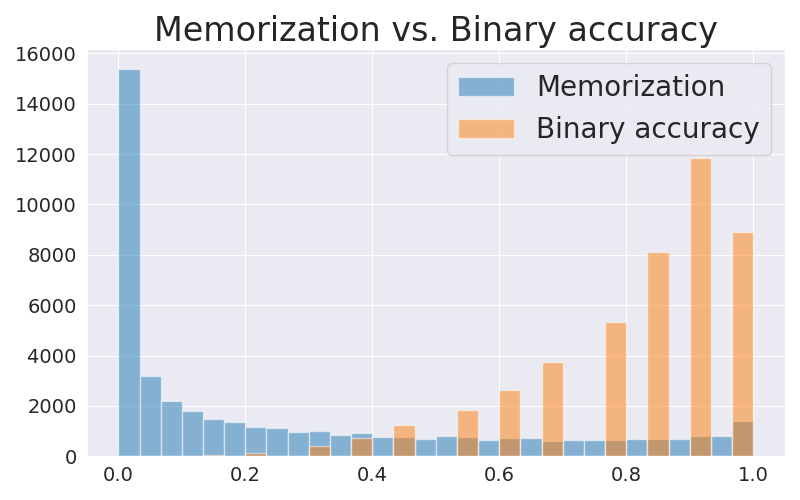}
% \caption{Tiny-ImageNet with VGG-16}
\end{subfigure}\begin{subfigure}[b]{0.32\textwidth}
\centering
\includegraphics[scale=0.2]{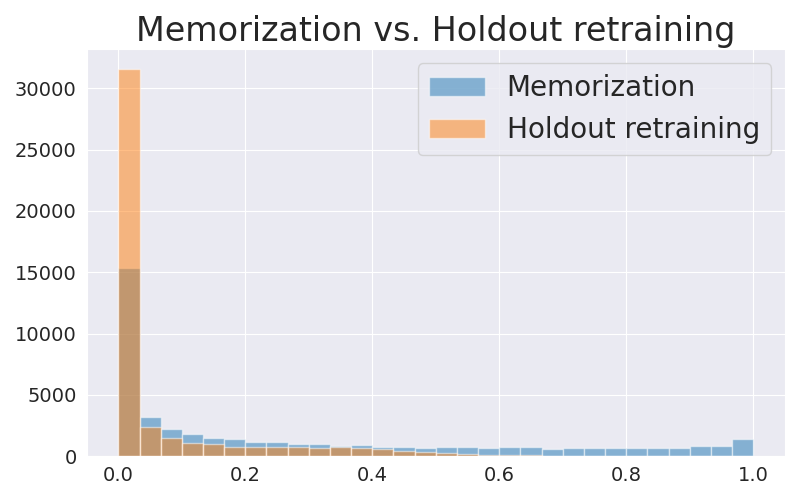}
% \caption{Tiny-ImageNet with VGG-16}
\end{subfigure}\begin{subfigure}[b]{0.32\textwidth}
\centering
\includegraphics[scale=0.2]{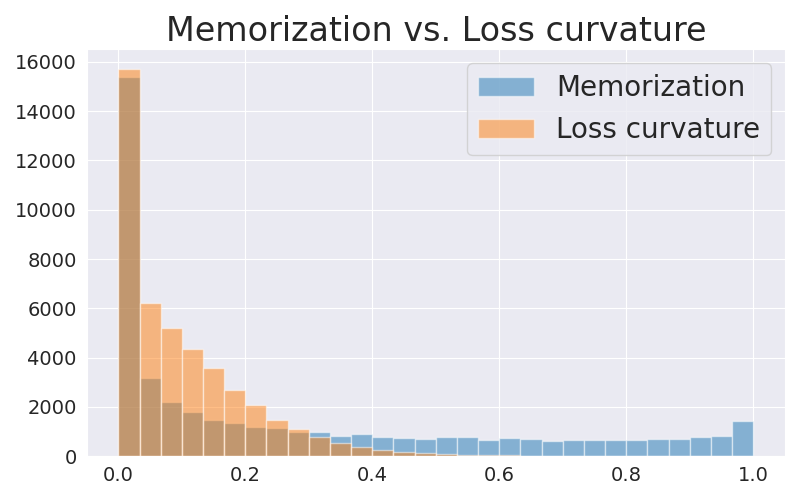}
% \caption{Tiny-ImageNet with VGG-16}
\end{subfigure}
\caption*{CIFAR-10 with ResNet-18} % Group caption
\end{minipage}

\vspace{1em} % Add vertical space between the groups

% Group 2: CIFAR-100
\begin{minipage}[b]{\textwidth}
\centering
\captionsetup{type=figure} % Allows for a group caption
\begin{subfigure}[b]{0.32\textwidth}
 \centering
\includegraphics[scale=0.2]{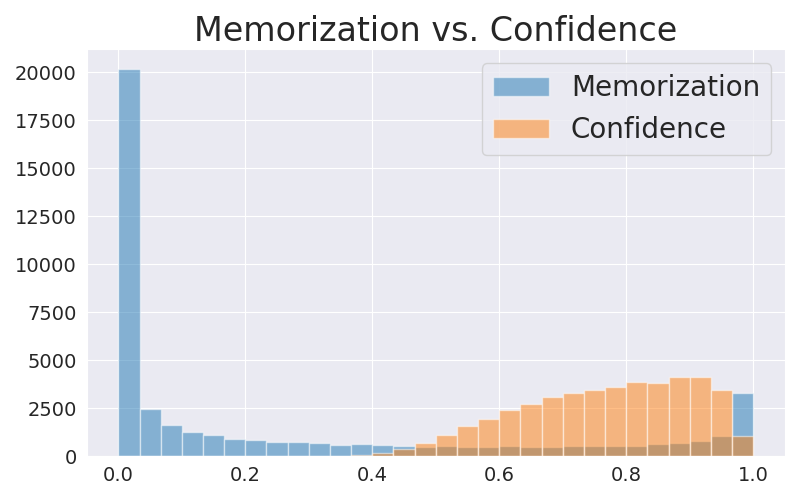}
 % \caption{Memorization vs Confidence}
\end{subfigure}
\begin{subfigure}[b]{0.32\textwidth}
\centering
\includegraphics[scale=0.2]{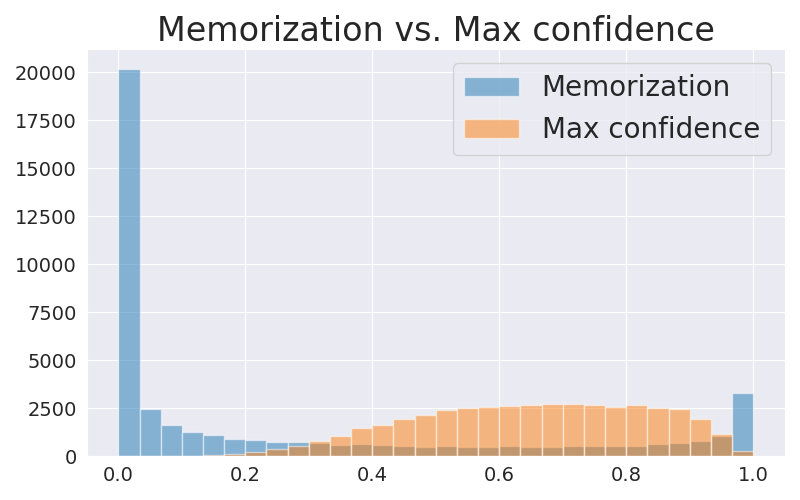}
 % \caption{Memorization vs Holdout retraining}
\end{subfigure}
\begin{subfigure}[b]{0.32\textwidth}
\centering
\includegraphics[scale=0.2]{figures/mem_entropy_hist_cifar10_resnet18_s1.png}
% \caption{Tiny-ImageNet with VGG-16}
\end{subfigure}
\begin{subfigure}[b]{0.32\textwidth}
\centering
\includegraphics[scale=0.2]{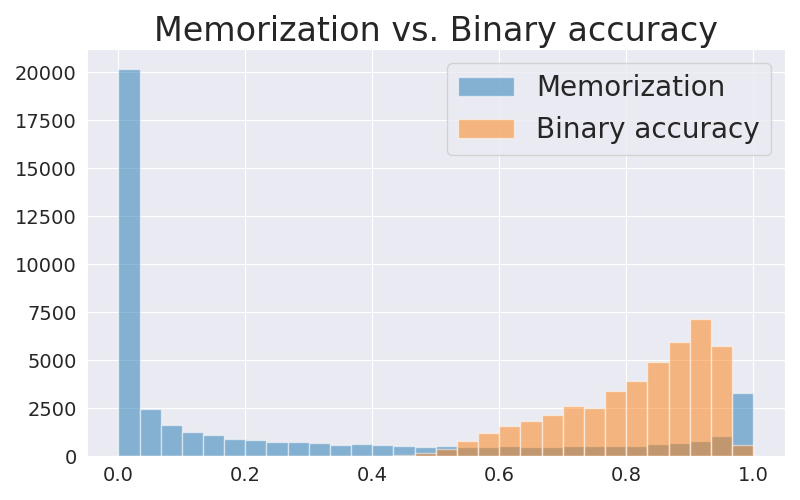}
% \caption{Tiny-ImageNet with VGG-16}
\end{subfigure}\begin{subfigure}[b]{0.32\textwidth}
\centering
\includegraphics[scale=0.2]{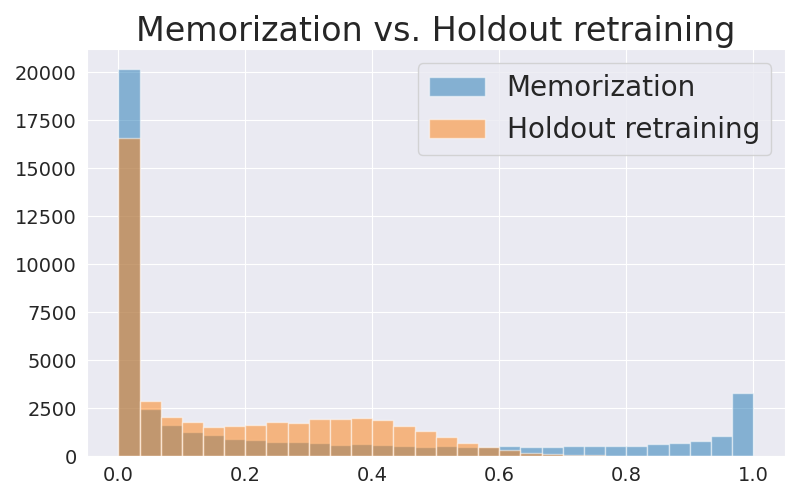}
% \caption{Tiny-ImageNet with VGG-16}
\end{subfigure}\begin{subfigure}[b]{0.32\textwidth}
\centering
\includegraphics[scale=0.2]{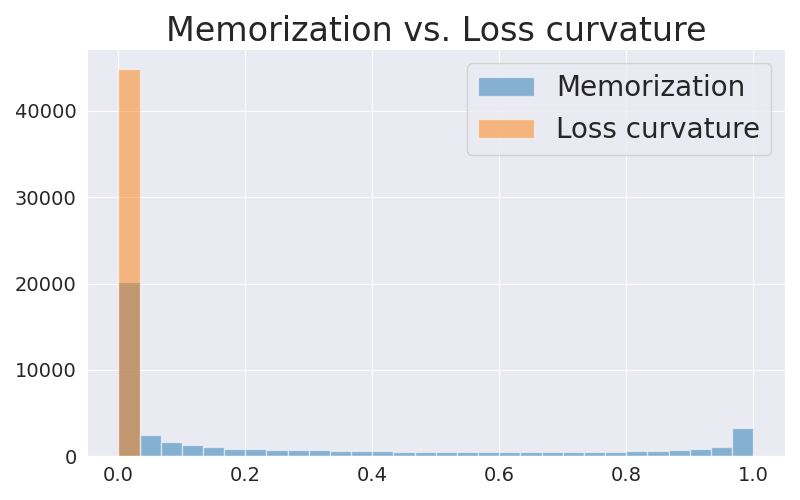}
% \caption{Tiny-ImageNet with VGG-16}
\end{subfigure}
\caption*{CIFAR-100 with ResNet-50} % Group caption
\end{minipage}
\caption{Distribution of memorization scores and proxies. Each plot compares a proxy with respect to the memorization scores. Results are presented for CIFAR-10 using ResNet-18 and CIFAR-100 using ResNet-50.}
\label{fig:proxy-distro}
\end{figure}

To illustrate the fidelity of each proxy in relation to memorization, we present a distribution comparison between memorization and each proxy. 
Figure \ref{fig:proxy-distro} displays these distributions, showing how each proxy relates to memorization. 
It is important to note that the learning event proxies (i.e., confidence, max confidence, entropy, and binary accuracy) are negatively correlated with memorization, as indicated by the Spearman correlation coefficients in Table \ref{tab:proxy-comparison}. This negative correlation is also evident in the distribution plots.

\subsubsection{RUM$^\mathcal{F}$ and control experiment results}
\label{sec:rum-results}

Table \ref{tab:tow-towmia-rte-full} and Table \ref{tab:acc-full} present the results of RUM$^\mathcal{F}$ and the corresponding control experiment, including "vanilla" and "shuffle", the description of which can be found in Section \ref{sec:rum-experiment}.
For each unlearning algorithm, we conduct three experiments: RUM$^\mathcal{F}$, shuffle and vanilla, and collect the forget, retain, and test accuracy, as well as MIA scores. We then calculate ToW and ToW-MIA using accuracies (as outlined in Section \ref{sec:eval}) to evaluate their performance.
Additionally, Table \ref{tab:tow-towmia-rte-full} includes a "runtime" column to report the running time for each experiment, highlighting the efficiency of each method.
This procedure is repeated for each proxy (i.e. confidence, binary accuracy, and holdout retraining) across three dataset/architecture settings: CIFAR-10 with ResNet-18, CIFAR-100 with ResNet-50, and Tiny-ImageNet with VGG-16.

\begin{figure}[t]
\centering
% Group 1: CIFAR-10
\begin{minipage}[b]{\textwidth}
\centering
\captionsetup{type=figure} % Allows for a group caption
\begin{subfigure}[b]{0.24\textwidth}
 \centering
\includegraphics[scale=0.18]{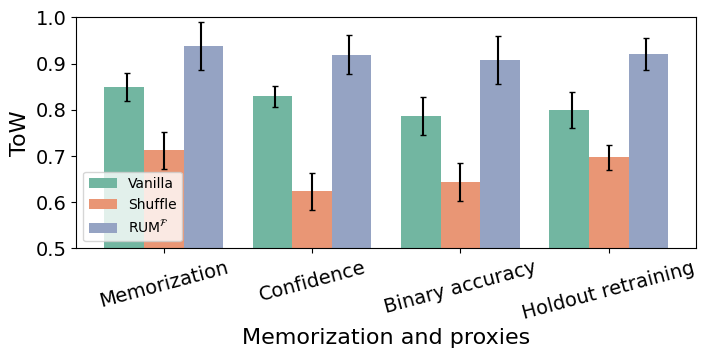}
 \caption{Fine-tune}
\end{subfigure}
\begin{subfigure}[b]{0.24\textwidth}
\centering
\includegraphics[scale=0.18]{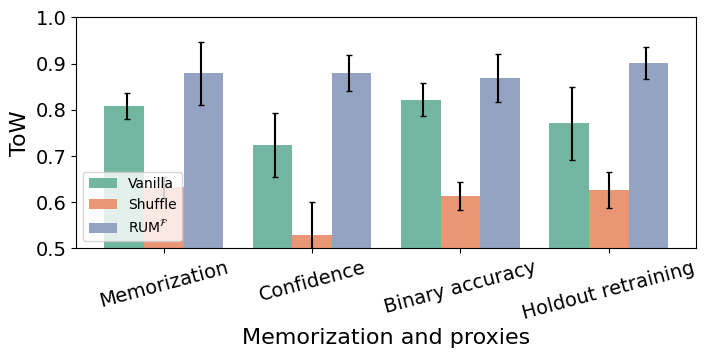}
 \caption{NegGrad+}
\end{subfigure}
\begin{subfigure}[b]{0.24\textwidth}
\centering
\includegraphics[scale=0.18]{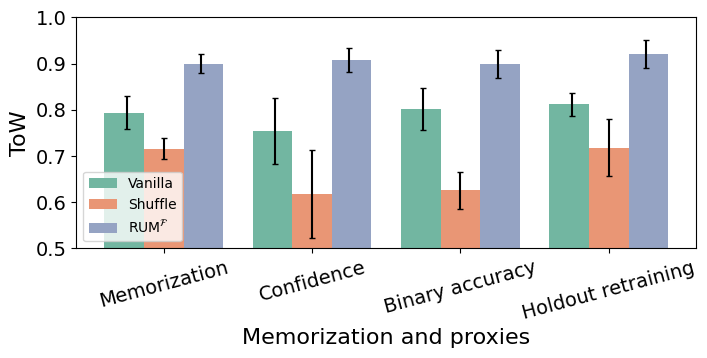}
\caption{L1-sparse}
\end{subfigure}
\begin{subfigure}[b]{0.24\textwidth}
\centering
\includegraphics[scale=0.18]{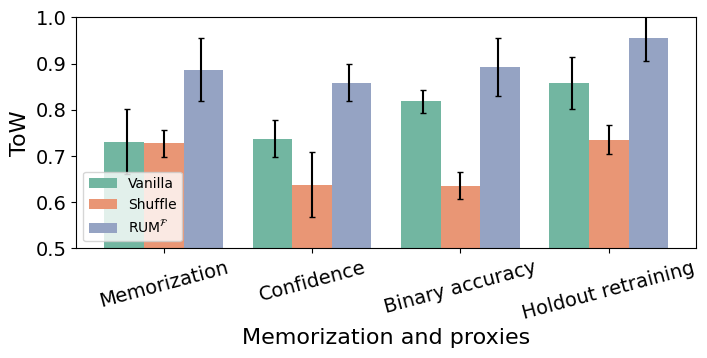}
\caption{SalUn}
\end{subfigure}
\caption*{CIFAR-10 with ResNet-18} % Group caption
\end{minipage}

\vspace{1em} % Add vertical space between the groups

% Group 2: CIFAR-100
\begin{minipage}[b]{\textwidth}
\centering
\captionsetup{type=figure} % Allows for a group caption
\begin{subfigure}[b]{0.24\textwidth}
 \centering
\includegraphics[scale=0.18]{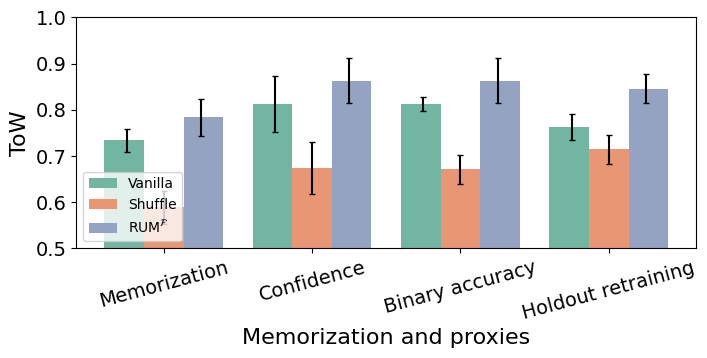}
 \caption{Fine-tune}
\end{subfigure}
\begin{subfigure}[b]{0.24\textwidth}
\centering
\includegraphics[scale=0.18]{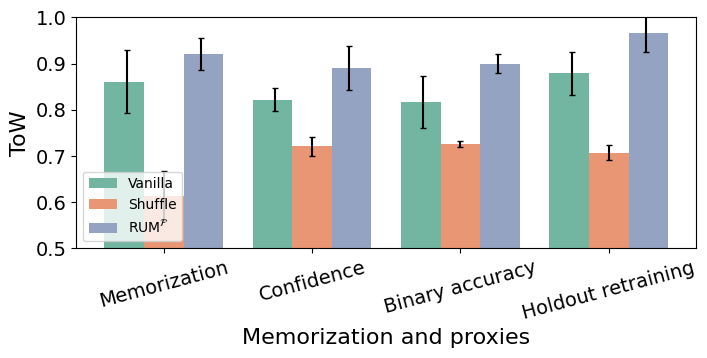}
 \caption{NegGrad+}
\end{subfigure}
\begin{subfigure}[b]{0.24\textwidth}
\centering
\includegraphics[scale=0.18]{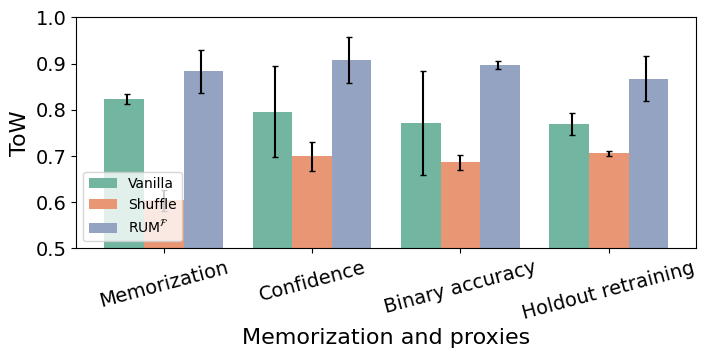}
\caption{L1-sparse}
\end{subfigure}
\begin{subfigure}[b]{0.24\textwidth}
\centering
\includegraphics[scale=0.18]{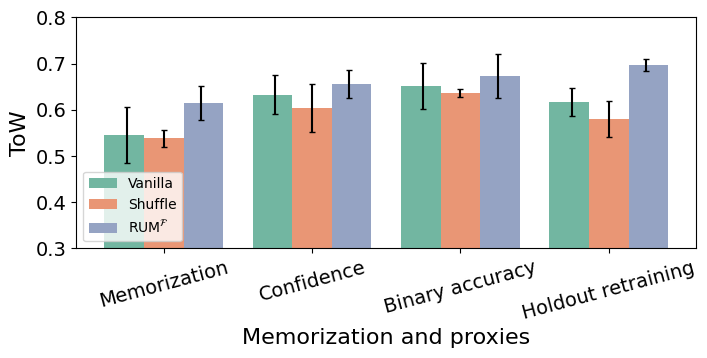}
\caption{SalUn}
\end{subfigure}
\caption*{CIFAR-100 with ResNet-50} % Group caption
\end{minipage}

\vspace{1em} % Add vertical space between the groups

% Group 3: Tiny imagenet
\begin{minipage}[b]{\textwidth}
\centering
\captionsetup{type=figure} % Allows for a group caption
\begin{subfigure}[b]{0.24\textwidth}
 \centering
\includegraphics[scale=0.18]{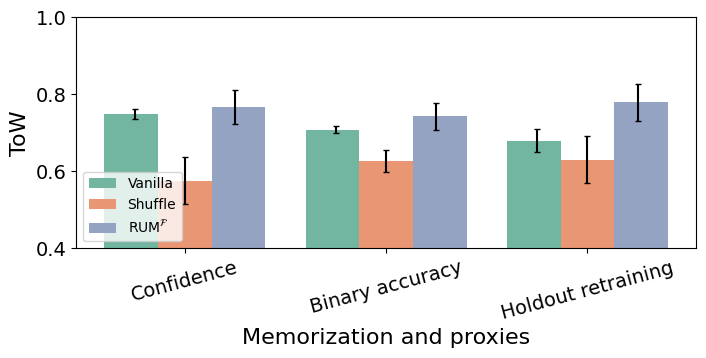}
 \caption{Fine-tune}
\end{subfigure}
\begin{subfigure}[b]{0.24\textwidth}
\centering
\includegraphics[scale=0.18]{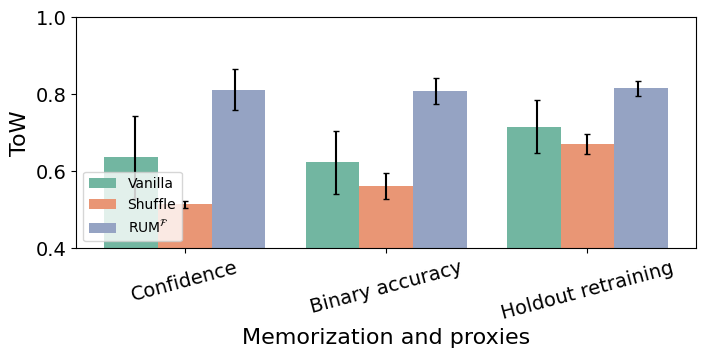}
 \caption{NegGrad+}
\end{subfigure}
\begin{subfigure}[b]{0.24\textwidth}
\centering
\includegraphics[scale=0.18]{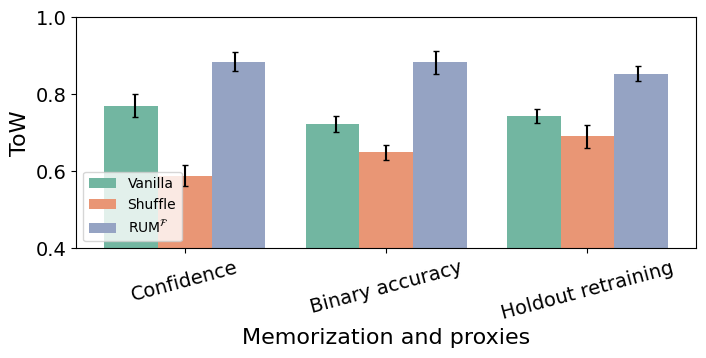}
\caption{L1-sparse}
\end{subfigure}
\begin{subfigure}[b]{0.24\textwidth}
\centering
\includegraphics[scale=0.18]{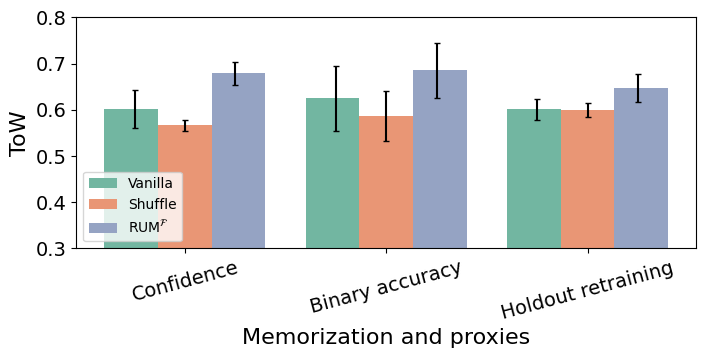}
\caption{SalUn}
\end{subfigure}
\caption*{Tiny-ImageNet with VGG-16} % Group caption
\end{minipage}

\caption{ToW results for RUM$^\mathcal{F}$, shuffle, and vanilla approaches across different proxies and memorization (for CIFAR-10 and CIFAR-100). For each unlearning algorithm (Fine-tune, NegGrad+, L1-sparse, and SalUn), we present ToW results across the three approaches (RUM$^\mathcal{F}$, shuffle, and vanilla) for each proxy and memorization (where applicable). The experiments were conducted on CIFAR-10 with ResNet-18, CIFAR-100 with ResNet-50, and Tiny-ImageNet with VGG-16, each repeated three times, with averages and 95\% confidence intervals reported.}
\label{fig:tow-full}
\end{figure}

\begin{figure}[t]
\centering
% Group 1: CIFAR-10
\begin{minipage}[b]{\textwidth}
\centering
\captionsetup{type=figure} % Allows for a group caption
\begin{subfigure}[b]{0.24\textwidth}
 \centering
\includegraphics[scale=0.18]{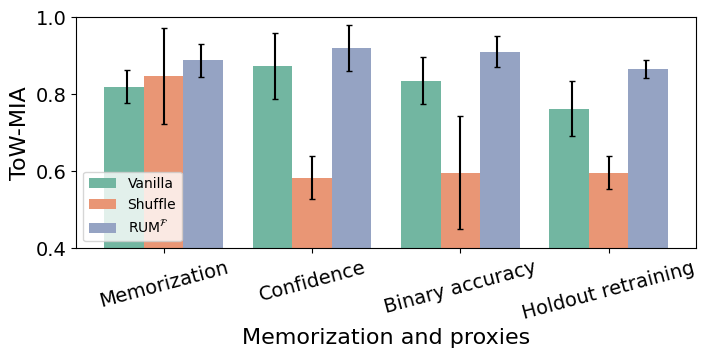}
 \caption{Fine-tune}
\end{subfigure}
\begin{subfigure}[b]{0.24\textwidth}
\centering
\includegraphics[scale=0.18]{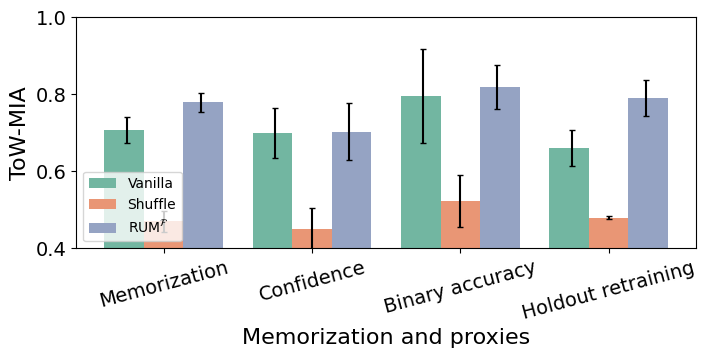}
 \caption{NegGrad+}
\end{subfigure}
\begin{subfigure}[b]{0.24\textwidth}
\centering
\includegraphics[scale=0.18]{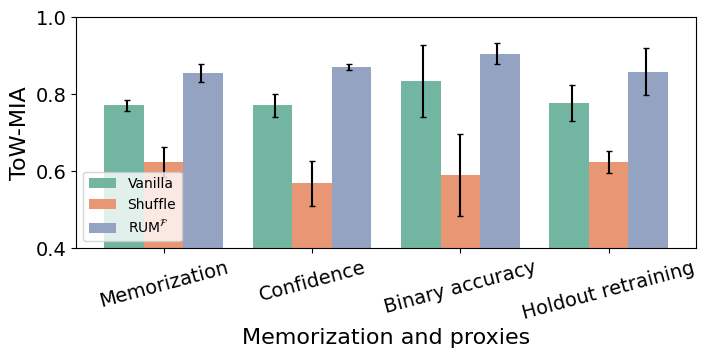}
\caption{L1-sparse}
\end{subfigure}
\begin{subfigure}[b]{0.24\textwidth}
\centering
\includegraphics[scale=0.18]{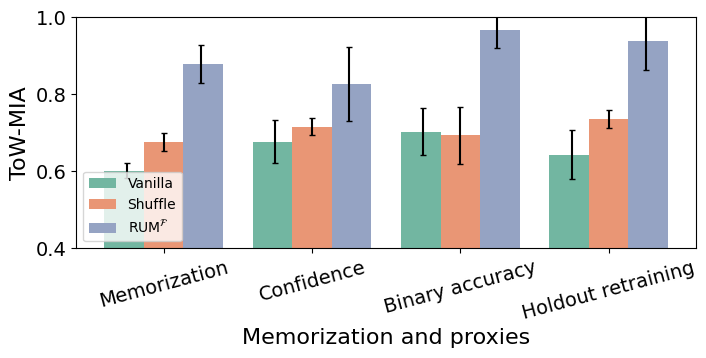}
\caption{SalUn}
\end{subfigure}
\caption*{CIFAR-10 with ResNet-18} % Group caption
\end{minipage}

\vspace{1em} % Add vertical space between the groups

% Group 2: CIFAR-100
\begin{minipage}[b]{\textwidth}
\centering
\captionsetup{type=figure} % Allows for a group caption
\begin{subfigure}[b]{0.24\textwidth}
 \centering
\includegraphics[scale=0.18]{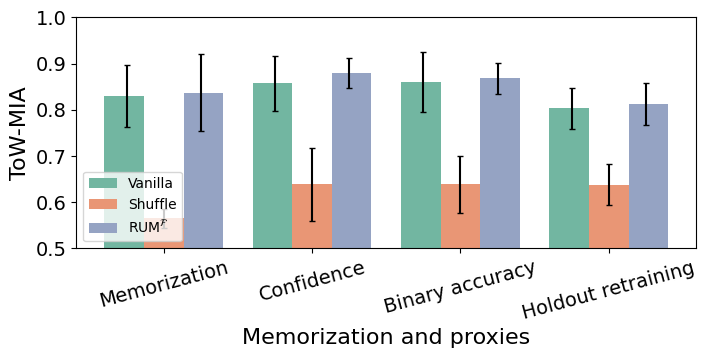}
 \caption{Fine-tune}
\end{subfigure}
\begin{subfigure}[b]{0.24\textwidth}
\centering
\includegraphics[scale=0.18]{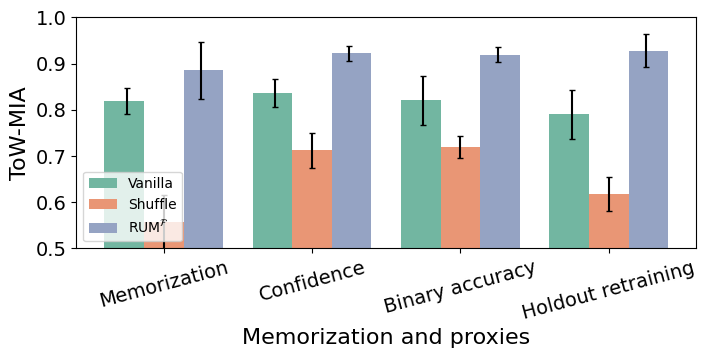}
 \caption{NegGrad+}
\end{subfigure}
\begin{subfigure}[b]{0.24\textwidth}
\centering
\includegraphics[scale=0.18]{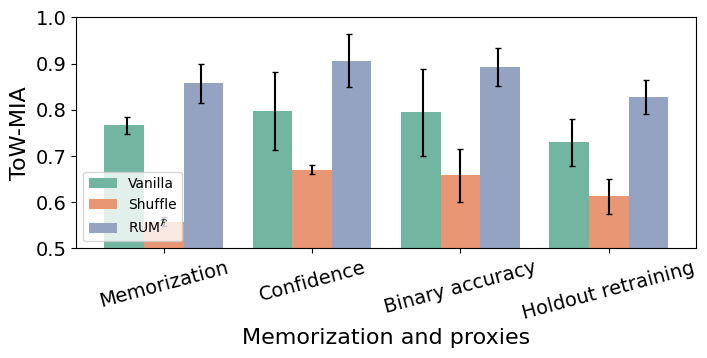}
\caption{L1-sparse}
\end{subfigure}
\begin{subfigure}[b]{0.24\textwidth}
\centering
\includegraphics[scale=0.18]{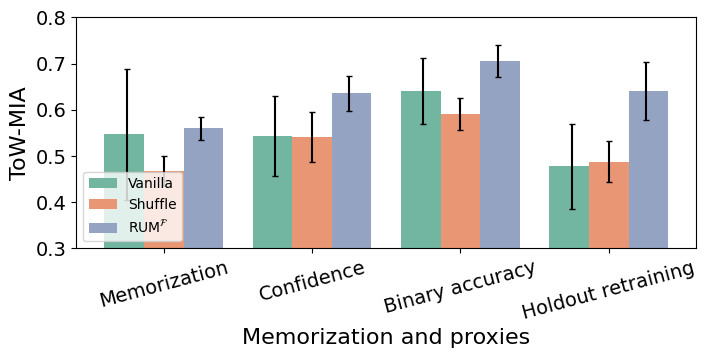}
\caption{SalUn}
\end{subfigure}
\caption*{CIFAR-100 with ResNet-50} % Group caption
\end{minipage}

\vspace{1em} % Add vertical space between the groups

% Group 3: Tiny imagenet
\begin{minipage}[b]{\textwidth}
\centering
\captionsetup{type=figure} % Allows for a group caption
\begin{subfigure}[b]{0.24\textwidth}
 \centering
\includegraphics[scale=0.18]{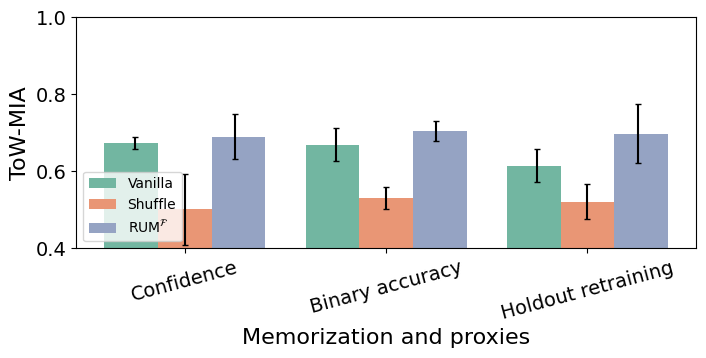}
 \caption{Fine-tune}
\end{subfigure}
\begin{subfigure}[b]{0.24\textwidth}
\centering
\includegraphics[scale=0.18]{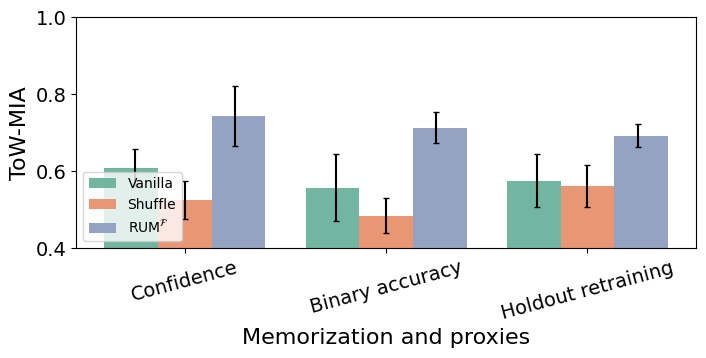}
 \caption{NegGrad+}
\end{subfigure}
\begin{subfigure}[b]{0.24\textwidth}
\centering
\includegraphics[scale=0.18]{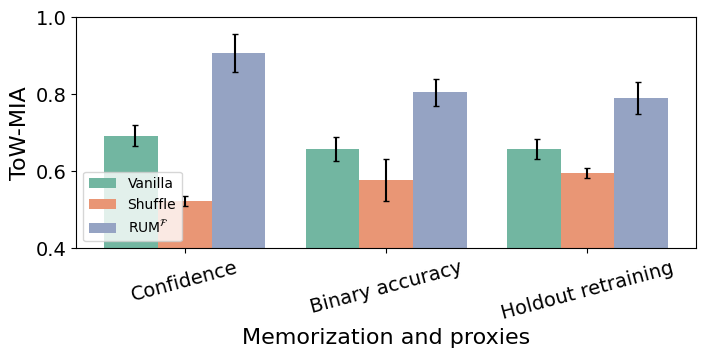}
\caption{L1-sparse}
\end{subfigure}
\begin{subfigure}[b]{0.24\textwidth}
\centering
\includegraphics[scale=0.18]{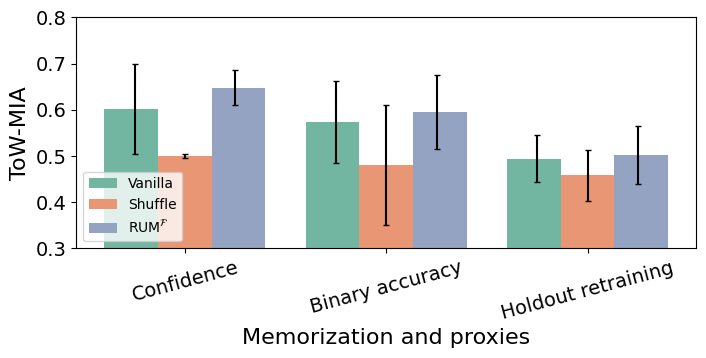}
\caption{SalUn}
\end{subfigure}
\caption*{Tiny-ImageNet with VGG-16} % Group caption
\end{minipage}

\caption{ToW-MIA results for RUM$^\mathcal{F}$, shuffle, and vanilla approaches using various proxies and memorization (for CIFAR-10 and CIFAR-100). For each unlearning algorithm (Fine-tune, NegGrad+, L1-sparse, and SalUn), ToW-MIA results are presented across the three approaches (RUM$^\mathcal{F}$, shuffle, and vanilla) for each proxy and memorization (where applicable). The experiments were performed on CIFAR-10 with ResNet-18, CIFAR-100 with ResNet-50, and Tiny-ImageNet with VGG-16, each run three times, with results reported as averages and 95\% confidence intervals.}
\label{fig:towmia-full}
\end{figure}

Figures \ref{fig:tow-full} and \ref{fig:towmia-full} provide visualizations of Table \ref{tab:tow-towmia-rte-full} in terms of ToW and ToW-MIA, respectively. 
Figure \ref{fig:tow-full} displays ToW, while Figure \ref{fig:towmia-full} illustrates ToW-MIA for the unlearning algorithms using different proxies (and memorization where applicable for CIFAR-10 and CIFAR-100) across various dataset/architecture settings.

\begin{table}[t]
\centering
\caption{Performance and runtime comparison of unlearning algorithms using confidence, binary accuracy, and holdout retraining proxies.
Each algorithm $\mathcal{U}$ is applied in three different approaches: i) in one go ("vanilla"), ii) sequentially on a random partition of $\forgetset$ into three equal-sized subsets ("shuffle"), and iii) sequentially on three equal-sized subsets obtained by $\refine$, processed in a low → medium → high memorization order based on the proxy ("RUM$^\mathcal{F}$").
Each experiment is repeated three times, and results are reported as averages with 95\% confidence intervals.}
\begin{subtable}{\textwidth}
\centering
\resizebox{\textwidth}{!}{
\begin{tabular}{lccc|ccc|ccc}
\toprule
\cmidrule(r){1-10}
& \multicolumn{3}{c}{Confidence} & \multicolumn{3}{c}{Binary accuracy} & \multicolumn{3}{c}{Holdout retraining} \\
{} & ToW ($\uparrow$) & ToW-MIA ($\uparrow$) & Runtime (s) & ToW ($\uparrow$) & ToW-MIA ($\uparrow$) & Runtime (s) & ToW ($\uparrow$) & ToW-MIA ($\uparrow$) & Runtime (s) \\
\midrule
\midrule
Retrain & 1.000 ± 0.000 & 1.000 ± 0.000 & {427.314} & 1.000 ± 0.000 & 1.000 ± 0.000 & 427.277 & 1.000 ± 0.000 & 1.000 ± 0.000 & 430.526 \\
\midrule
Fine-tune RUM$^\mathcal{F}$ & 0.919 ± 0.042 & 0.920 ± 0.059 & {288.163} & 0.908 ± 0.052 & 0.911 ± 0.041 & 280.439 & 0.920 ± 0.035 & 0.865 ± 0.023 & 320.511 \\
Fine-tune shuffle & 0.624 ± 0.040 & 0.584 ± 0.056 & {291.465} & 0.644 ± 0.041 & 0.597 ± 0.147 & 295.766 & 0.697 ± 0.027 & 0.597 ± 0.043 & 336.668 \\
Fine-tune vanilla & 0.829 ± 0.022 & 0.874 ± 0.185 & {166.337} & 0.787 ± 0.041 & 0.836 ± 0.062 & 163.180 & 0.800 ± 0.039 & 0.763 ± 0.072 & 205.539 \\
\midrule
NegGrad+ RUM$^\mathcal{F}$ & 0.880 ± 0.039 & 0.803 ± 0.074 & {376.308} & 0.868 ± 0.052 & 0.819 ± 0.057 & 375.025 & 0.901 ± 0.035 & 0.791 ± 0.046 & 372.202 \\
NegGrad+ shuffle & 0.529 ± 0.071 & 0.450 ± 0.056 & {378.119} & 0.613 ± 0.031 & 0.523 ± 0.068 & 382.487 & 0.626 ± 0.039 & 0.480 ± 0.040 & 384.028 \\
NegGrad+ vanilla & 0.724 ± 0.070 & 0.700 ± 0.065 & {167.852} & 0.822 ± 0.036 & 0.795 ± 0.122 & 163.180 & 0.771 ± 0.079 & 0.661 ± 0.047 & 140.775 \\
\midrule
L1-sparse RUM$^\mathcal{F}$ & 0.907 ± 0.026 & 0.870 ± 0.008 & {285.494} & 0.899 ± 0.030 & 0.906 ± 0.027 & 303.667 & 0.920 ± 0.030 & 0.859 ± 0.061 & 291.099 \\
L1-sparse shuffle & 0.618 ± 0.095 & 0.569 ± 0.059 & {283.997} & 0.626 ± 0.040 & 0.591 ± 0.107 & 297.092 & 0.718 ± 0.062 & 0.624 ± 0.062 & 280.279 \\
L1-sparse vanilla & 0.754 ± 0.071 & 0.772 ± 0.030 & {156.423} & 0.802 ± 0.046 & 0.834 ± 0.094 & 154.346 & 0.812 ± 0.025 & 0.778 ± 0.047 & 158.543 \\
\midrule
SalUn RUM$^\mathcal{F}$ & 0.859 ± 0.042 & 0.867 ± 0.097 & {234.165} & 0.892 ± 0.063 & 0.966 ± 0.045 & 227.167 & 0.956 ± 0.051 & 0.939 ± 0.076 & 230.375 \\
SalUn shuffle & 0.638 ± 0.071 & 0.716 ± 0.022 & {240.078} & 0.636 ± 0.030 & 0.694 ± 0.074 & 227.042 & 0.735 ± 0.031 & 0.736 ± 0.024 & 226.220 \\
SalUn vanilla & 0.737 ± 0.040 & 0.677 ± 0.056 & {90.334} & 0.818 ± 0.024 & 0.703 ± 0.061 & 86.423 & 0.858 ± 0.056 & 0.643 ± 0.064 & 92.601 \\
\bottomrule
\end{tabular}}
\caption{CIFAR-10 with ResNet-18}
\end{subtable}

\begin{subtable}{\textwidth}
\centering
\resizebox{\textwidth}{!}{
\begin{tabular}{lccc|ccc|ccc}
\toprule
\cmidrule(r){1-10}
& \multicolumn{3}{c}{Confidence} & \multicolumn{3}{c}{Binary accuracy} & \multicolumn{3}{c}{Holdout retraining} \\
{} & ToW ($\uparrow$) & ToW-MIA ($\uparrow$) & Runtime (s) & ToW ($\uparrow$) & ToW-MIA ($\uparrow$) & Runtime (s) & ToW ($\uparrow$) & ToW-MIA ($\uparrow$) & Runtime (s) \\
\midrule
\midrule
Retrain & 1.000 ± 0.000 & 1.000 ± 0.000 & {6254.604} & 1.000 ± 0.000 & 1.000 ± 0.000 & 6127.849 & 1.000 ± 0.000 & 1.000 ± 0.000 & 6430.247 \\
\midrule
Fine-tune RUM$^\mathcal{F}$ & 0.863 ± 0.049 & 0.857 ± 0.059 & 852.886 & 0.863 ± 0.048 & 0.861 ± 0.065 & 859.824 & 0.846 ± 0.032 & 0.812 ± 0.146 & 798.843 \\
Fine-tune shuffle & 0.674 ± 0.057 & 0.639 ± 0.079 & 843.585 & 0.671 ± 0.031 & 0.639 ± 0.062 & 865.150 & 0.714 ± 0.031 & 0.638 ± 0.045 & 852.442 \\
Fine-tune vanilla & 0.813 ± 0.061 & 0.880 ± 0.032 & 379.337 & 0.813 ± 0.015 & 0.868 ± 0.034 & 390.707 & 0.763 ± 0.028 & 0.803 ± 0.044 & 432.678 \\
\midrule
NegGrad+ RUM$^\mathcal{F}$ & 0.890 ± 0.047 & 0.922 ± 0.017 & 773.603 & 0.900 ± 0.020 & 0.919 ± 0.016 & 768.227 & 0.966 ± 0.042 & 0.928 ± 0.035 & 777.204 \\
NegGrad+ shuffle & 0.721 ± 0.020 & 0.712 ± 0.038 & 773.607 & 0.726 ± 0.007 & 0.719 ± 0.024 & 769.536 & 0.707 ± 0.016 & 0.618 ± 0.037 & 770.538 \\
NegGrad+ vanilla & 0.822 ± 0.025 & 0.836 ± 0.030 & 369.705 & 0.817 ± 0.053 & 0.821 ± 0.053 & 363.956 & 0.879 ± 0.046 & 0.790 ± 0.053 & 357.868 \\
\midrule
L1-sparse RUM$^\mathcal{F}$ & 0.908 ± 0.049 & 0.906 ± 0.057 & 783.477 & 0.897 ± 0.009 & 0.892 ± 0.041 & 782.627 & 0.867 ± 0.049 & 0.828 ± 0.037 & 769.910 \\
L1-sparse shuffle & 0.699 ± 0.031 & 0.670 ± 0.010 & 787.643 & 0.686 ± 0.016 & 0.658 ± 0.057 & 785.941 & 0.706 ± 0.005 & 0.613 ± 0.038 & 783.263 \\
L1-sparse vanilla & 0.796 ± 0.099 & 0.797 ± 0.084 & 395.368 & 0.771 ± 0.112 & 0.795 ± 0.094 & 396.259 & 0.770 ± 0.024 & 0.730 ± 0.115 & 397.543 \\
\midrule
SalUn RUM$^\mathcal{F}$ & 0.656 ± 0.031 & 0.636 ± 0.038 & 791.166 & 0.673 ± 0.048 & 0.641 ± 0.072 & 793.327 & 0.696 ± 0.013 & 0.640 ± 0.129 & 793.793 \\
SalUn shuffle & 0.603 ± 0.052 & 0.541 ± 0.055 & 792.552 & 0.636 ± 0.030 & 0.591 ± 0.035 & 795.967 & 0.581 ± 0.039 & 0.488 ± 0.045 & 793.672 \\
SalUn vanilla & 0.633 ± 0.043 & 0.543 ± 0.186 & 417.232 & 0.651 ± 0.050 & 0.705 ± 0.035 & 421.418 & 0.617 ± 0.030 & 0.478 ± 0.163 & 396.784 \\
\bottomrule
\end{tabular}}
\caption{CIFAR-100 with ResNet-50}
\end{subtable}

\begin{subtable}{\textwidth}
\centering
\resizebox{\textwidth}{!}{
\begin{tabular}{lccc|ccc|ccc}
\toprule
\cmidrule(r){1-10}
& \multicolumn{3}{c}{Confidence} & \multicolumn{3}{c}{Binary accuracy} & \multicolumn{3}{c}{Holdout retraining} \\
{} & ToW ($\uparrow$) & ToW-MIA ($\uparrow$) & Runtime (s) & ToW ($\uparrow$) & ToW-MIA ($\uparrow$) & Runtime (s) & ToW ($\uparrow$) & ToW-MIA ($\uparrow$) & Runtime (s) \\
\midrule
\midrule
Retrain & 1.000 ± 0.000 & 1.000 ± 0.000 & 5127.197 & 1.000 ± 0.000 & 1.000 ± 0.000 & 4952.773 & 1.000 ± 0.000 & 1.000 ± 0.000 & 5544.903 \\
\midrule
Fine-tune RUM$^\mathcal{F}$ & 0.812 ± 0.053 & 0.690 ± 0.058 & 1015.630 & 0.743 ± 0.035 & 0.705 ± 0.026 & 1056.892 & 0.779 ± 0.047 & 0.698 ± 0.076 & 1020.360 \\
Fine-tune shuffle & 0.514 ± 0.009 & 0.501 ± 0.091 & 1031.002 & 0.627 ± 0.029 & 0.531 ± 0.028 & 1022.625 & 0.630 ± 0.061 & 0.521 ± 0.046 & 1023.901 \\
Fine-tune vanilla & 0.637 ± 0.106 & 0.673 ± 0.015 & 496.984 & 0.708 ± 0.009 & 0.669 ± 0.043 & 499.738 & 0.679 ± 0.030 & 0.615 ± 0.043 & 495.645 \\
\midrule
NegGrad+ RUM$^\mathcal{F}$ & 0.885 ± 0.024 & 0.743 ± 0.078 & 843.334 & 0.808 ± 0.034 & 0.713 ± 0.040 & 847.294 & 0.816 ± 0.020 & 0.692 ± 0.030 & 846.290 \\
NegGrad+ shuffle & 0.589 ± 0.028 & 0.526 ± 0.150 & 848.724 & 0.563 ± 0.034 & 0.485 ± 0.045 & 846.873 & 0.671 ± 0.025 & 0.562 ± 0.055 & 845.253 \\
NegGrad+ vanilla & 0.771 ± 0.030 & 0.609 ± 0.048 & 488.123 & 0.624 ± 0.082 & 0.558 ± 0.087 & 491.732 & 0.716 ± 0.069 & 0.576 ± 0.069 & 483.439 \\
\midrule
L1-sparse RUM$^\mathcal{F}$ & 0.767 ± 0.044 & 0.907 ± 0.050 & 806.103 & 0.883 ± 0.050 & 0.806 ± 0.043 & 816.238 & 0.854 ± 0.019 & 0.791 ± 0.042 & 806.247 \\
L1-sparse shuffle & 0.576 ± 0.061 & 0.523 ± 0.013 & 818.745 & 0.649 ± 0.019 & 0.578 ± 0.055 & 806.770 & 0.691 ± 0.030 & 0.596 ± 0.012 & 814.305 \\
L1-sparse vanilla & 0.750 ± 0.013 & 0.693 ± 0.028 & 498.304 & 0.723 ± 0.020 & 0.658 ± 0.031 & 507.753 & 0.744 ± 0.018 & 0.657 ± 0.026 & 507.526 \\
\midrule
SalUn RUM$^\mathcal{F}$ & 0.679 ± 0.025 & 0.602 ± 0.097 & 832.785 & 0.685 ± 0.059 & 0.595 ± 0.054 & 835.974 & 0.647 ± 0.030 & 0.502 ± 0.063 & 833.936 \\
SalUn shuffle & 0.566 ± 0.011 & 0.500 ± 0.005 & 835.109 & 0.587 ± 0.054 & 0.481 ± 0.130 & 838.125 & 0.599 ± 0.015 & 0.458 ± 0.055 & 829.783 \\
SalUn vanilla & 0.602 ± 0.041 & 0.648 ± 0.037 & 483.636 & 0.625 ± 0.070 & 0.573 ± 0.189 & 491.136 & 0.601 ± 0.023 & 0.494 ± 0.051 & 481.330 \\
\bottomrule
\end{tabular}}
\caption{Tiny-ImageNet with VGG-16}
\end{subtable}
\label{tab:tow-towmia-rte-full}
\end{table}

\begin{table}[t]
\centering
\caption{Accuracy and MIA performance for different unlearning algorithms across various proxies on CIFAR-10/ResNet-18, CIFAR-100/ResNet-50, and Tiny-ImageNet/VGG-16. Results are averaged over 3 runs, with 95\% confidence intervals reported.
% \todo{double check the numbers}
}
\begin{subtable}{\textwidth}
\centering
\resizebox{\textwidth}{!}{
\begin{tabular}{lcccc|cccc|cccc}
\toprule
& \multicolumn{4}{c}{Confidence} & \multicolumn{4}{c}{Binary accuracy} & \multicolumn{4}{c}{Holdout retraining} \\
{} & Retain Acc & Forget Acc & Test Acc & MIA & Retain Acc & Forget Acc & Test Acc & MIA & Retain Acc & Forget Acc & Test Acc & MIA \\
\midrule
\midrule
Retrain & 100.000 ± 0.000  & 50.433 ± 6.808  & 84.167 ± 1.616  & 0.637 ± 0.032 & 100.000 ± 0.000  & 47.156 ± 9.892  & 83.683 ± 0.953  & 0.663 ± 0.038  & 100.000 ± 0.000  & 62.922 ± 4.681  & 84.270 ± 2.183  & 0.564 ± 0.022 \\
\midrule
Fine-tune RUM$^\mathcal{F}$ & 99.460 ± 1.419  & 56.189 ± 5.483  & 82.320 ± 5.818  & 0.579 ± 0.039 & 99.207 ± 2.186  & 54.333 ± 10.423  & 82.323 ± 2.336  & 0.594 ± 0.052  & 98.157 ± 1.818  & 66.256 ± 4.692  & 81.270 ± 3.304  & 0.472 ± 0.004 \\
Fine-tune shuffle     & 95.491 ± 4.235  & 82.089 ± 6.515  & 79.940 ± 8.227  & 0.276 ± 0.077 & 96.778 ± 4.003  & 79.578 ± 18.263  & 82.290 ± 4.460  & 0.291 ± 0.242  & 95.275 ± 6.451  & 86.178 ± 4.784  & 79.697 ± 10.533 & 0.221 ± 0.077 \\
Fine-tune vanilla     & 98.060 ± 5.309  & 62.800 ± 16.311 & 80.670 ± 5.082  & 0.560 ± 0.091 & 98.612 ± 2.610  & 64.867 ± 9.118   & 80.670 ± 6.397  & 0.537 ± 0.059  & 97.967 ± 3.280  & 77.244 ± 6.718  & 79.647 ± 1.761  & 0.380 ± 0.009 \\
\midrule
NegGrad+ RUM$^\mathcal{F}$ & 99.037 ± 1.532  & 56.067 ± 5.261  & 78.443 ± 5.531  & 0.504 ± 0.049 & 98.696 ± 1.376  & 48.322 ± 6.839   & 75.883 ± 3.480  & 0.564 ± 0.049  & 98.748 ± 1.310  & 64.089 ± 8.840  & 76.627 ± 4.501  & 0.432 ± 0.066 \\
NegGrad+ shuffle & 98.919 ± 2.120  & 94.867 ± 9.081  & 80.563 ± 6.630  & 0.110 ± 0.107 & 93.426 ± 6.033  & 75.189 ± 17.793  & 74.953 ± 5.193  & 0.278 ± 0.155  & 99.612 ± 0.331  & 97.667 ± 2.518  & 80.633 ± 1.315  & 0.064 ± 0.031 \\
NegGrad+ vanilla & 91.123 ± 12.963 & 56.300 ± 37.655 & 73.510 ± 10.680 & 0.503 ± 0.301 & 95.580 ± 2.430  & 51.511 ± 21.793  & 76.937 ± 3.304  & 0.556 ± 0.151  & 97.534 ± 6.008  & 78.478 ± 3.000  & 77.867 ± 11.231 & 0.289 ± 0.043 \\
\midrule
L1-sparse RUM$^\mathcal{F}$ & 96.947 ± 1.991  & 53.611 ± 3.446  & 80.783 ± 2.540  & 0.566 ± 0.021 & 99.190 ± 0.443  & 55.244 ± 8.641   & 82.347 ± 2.069  & 0.589 ± 0.042  & 97.694 ± 3.078  & 65.600 ± 4.265  & 81.110 ± 2.920  & 0.472 ± 0.019 \\
L1-sparse shuffle & 95.890 ± 4.480  & 83.222 ± 1.391  & 79.963 ± 5.127  & 0.256 ± 0.031 & 96.273 ± 3.580  & 80.622 ± 14.632  & 81.493 ± 4.575  & 0.291 ± 0.178  & 95.832 ± 4.322  & 85.322 ± 4.436  & 80.807 ± 2.414  & 0.239 ± 0.053 \\
L1-sparse vanilla & 96.095 ± 4.676  & 66.856 ± 9.041  & 78.200 ± 5.173  & 0.492 ± 0.041 & 97.175 ± 6.035  & 60.811 ± 14.615  & 79.323 ± 3.242  & 0.560 ± 0.065  & 96.794 ± 6.533  & 74.500 ± 6.474  & 79.320 ± 0.410  & 0.410 ± 0.057 \\
\midrule
SalUn RUM$^\mathcal{F}$ & 98.030 ± 1.356  & 58.111 ± 4.816  & 79.180 ± 4.306  & 0.593 ± 0.014 & 99.891 ± 0.085  & 56.400 ± 6.402   & 82.303 ± 5.425  & 0.645 ± 0.015  & 99.757 ± 0.651  & 61.500 ± 5.984  & 82.683 ± 3.331  & 0.607 ± 0.039 \\
SalUn shuffle  & 97.897 ± 0.449  & 84.067 ± 5.039  & 82.387 ± 3.331  & 0.381 ± 0.015 & 97.823 ± 0.409  & 81.189 ± 7.998   & 82.367 ± 4.221  & 0.384 ± 0.091  & 97.902 ± 0.325  & 86.444 ± 3.850  & 82.447 ± 4.513  & 0.330 ± 0.023 \\
SalUn vanilla  & 99.996 ± 0.009  & 75.578 ± 4.606  & 82.597 ± 4.474  & 0.949 ± 0.049 & 99.998 ± 0.007  & 63.878 ± 8.252   & 82.030 ± 4.742  & 0.948 ± 0.017  & 99.991 ± 0.021  & 75.389 ± 6.614  & 82.333 ± 4.831  & 0.909 ± 0.044 \\
\bottomrule
\end{tabular}}
\caption{CIFAR-10 with ResNet-18}
\end{subtable}

\begin{subtable}{\textwidth}
\centering
\resizebox{\textwidth}{!}{
\begin{tabular}{lcccc|cccc|cccc}
\toprule
& \multicolumn{4}{c}{Confidence} & \multicolumn{4}{c}{Binary accuracy} & \multicolumn{4}{c}{Holdout retraining} \\
{} & Retain Acc & Forget Acc & Test Acc & MIA & Retain Acc & Forget Acc & Test Acc & MIA & Retain Acc & Forget Acc & Test Acc & MIA \\
\midrule
\midrule
Retrain & 99.994 ± 0.007  & 64.267 ± 0.504  & 74.160 ± 1.623  & 0.473 ± 0.015 & 99.997 ± 0.003  & 64.511 ± 0.621  & 75.153 ± 1.427  & 0.465 ± 0.019  & 99.963 ± 0.025  & 69.856 ± 2.620  & 74.030 ± 1.237  & 0.479 ± 0.060 \\
\midrule
FTine-tune RUM$^\mathcal{F}$ & 97.002 ± 1.970  & 66.811 ± 3.428  & 65.460 ± 3.965  & 0.441 ± 0.010 & 96.796 ± 2.717  & 64.867 ± 1.242  & 64.597 ± 4.174  & 0.459 ± 0.007  & 96.281 ± 6.211  & 72.722 ± 9.196  & 64.643 ± 3.474  & 0.408 ± 0.024 \\
Fine-tune shuffle & 93.547 ± 10.320  & 83.722 ± 9.360  & 63.767 ± 9.213  & 0.236 ± 0.066 & 94.649 ± 7.021  & 84.856 ± 7.825  & 64.243 ± 6.960  & 0.222 ± 0.054  & 95.825 ± 4.014  & 87.844 ± 5.113  & 64.883 ± 4.310  & 0.212 ± 0.041 \\
Fine-tune vanilla & 99.692 ± 0.536  & 80.467 ± 8.165  & 71.453 ± 4.990  & 0.380 ± 0.017 & 99.502 ± 0.702  & 78.689 ± 2.530  & 70.403 ± 4.854  & 0.381 ± 0.022  & 99.751 ± 0.248  & 91.400 ± 3.803  & 71.667 ± 4.725  & 0.306 ± 0.016 \\
\midrule
NegGrad+ RUM$^\mathcal{F}$ & 98.635 ± 1.294  & 58.811 ± 4.524  & 69.630 ± 1.635  & 0.478 ± 0.057 & 99.230 ± 0.525  & 60.522 ± 1.198  & 69.597 ± 2.596  & 0.445 ± 0.029  & 99.838 ± 0.261  & 69.633 ± 4.341  & 71.450 ± 3.253  & 0.433 ± 0.052 \\
NegGrad+ shuffle & 94.895 ± 4.593  & 81.344 ± 8.248  & 65.867 ± 4.353  & 0.293 ± 0.118 & 97.266 ± 0.806  & 84.989 ± 4.567  & 69.090 ± 4.641  & 0.253 ± 0.071  & 94.213 ± 1.751  & 87.744 ± 0.172  & 65.370 ± 2.599  & 0.198 ± 0.036 \\
NegGrad+ vanilla & 97.755 ± 1.210  & 54.233 ± 2.348  & 67.660 ± 1.851  & 0.558 ± 0.018 & 98.118 ± 1.033  & 54.167 ± 5.900  & 67.990 ± 3.576  & 0.564 ± 0.037  & 99.679 ± 0.885  & 79.144 ± 7.291  & 71.260 ± 2.346  & 0.294 ± 0.086 \\
\midrule
L1-sparse RUM$^\mathcal{F}$ & 98.621 ± 1.213  & 65.444 ± 2.082  & 67.370 ± 3.775  & 0.458 ± 0.012 & 98.187 ± 0.418  & 65.344 ± 2.036  & 67.277 ± 3.352  & 0.451 ± 0.027  & 96.182 ± 1.555  & 68.544 ± 1.081  & 65.280 ± 2.923  & 0.422 ± 0.037 \\
L1-sparse shuffle     & 96.825 ± 0.943  & 86.889 ± 1.130  & 67.493 ± 1.459  & 0.215 ± 0.018 & 95.427 ± 3.689  & 85.300 ± 5.243  & 65.980 ± 5.357  & 0.223 ± 0.016  & 93.512 ± 1.992  & 85.711 ± 3.398  & 63.780 ± 2.756  & 0.210 ± 0.027 \\
L1-sparse vanilla     & 93.144 ± 5.025  & 61.011 ± 1.284  & 62.443 ± 4.461  & 0.442 ± 0.007 & 92.363 ± 6.279  & 60.311 ± 3.562  & 62.253 ± 5.650  & 0.453 ± 0.016  & 96.879 ± 3.757  & 84.578 ± 7.648  & 67.240 ± 3.661  & 0.287 ± 0.038 \\
\midrule
SalUn RUM$^\mathcal{F}$ & 93.047 ± 9.588  & 85.933 ± 14.590  & 64.467 ± 6.889  & 0.232 ± 0.119 & 88.792 ± 4.571  & 76.667 ± 9.624  & 61.437 ± 1.364  & 0.302 ± 0.108  & 92.798 ± 15.973  & 87.167 ± 24.725  & 66.337 ± 18.105  & 0.233 ± 0.060 \\
SalUn shuffle  & 79.560 ± 10.370  & 72.278 ± 9.983  & 56.617 ± 5.759  & 0.299 ± 0.086 & 88.594 ± 4.984  & 81.389 ± 5.900  & 61.583 ± 2.134  & 0.237 ± 0.012  & 73.911 ± 5.088  & 69.867 ± 5.380  & 53.297 ± 0.989  & 0.312 ± 0.051 \\
SalUn vanilla  & 84.519 ± 20.376  & 76.856 ± 27.639  & 60.637 ± 4.232  & 0.410 ± 0.777 & 98.818 ± 2.541  & 94.833 ± 18.050  & 69.810 ± 0.447  & 0.219 ± 0.056  & 76.144 ± 6.933  & 73.500 ± 9.736  & 58.250 ± 0.568  & 0.469 ± 0.767 \\
\bottomrule
\end{tabular}}
\caption{CIFAR-100 with ResNet-50}
\end{subtable}

\begin{subtable}{\textwidth}
\centering
\resizebox{\textwidth}{!}{
\begin{tabular}{lcccc|cccc|cccc}
\toprule
& \multicolumn{4}{c}{Confidence} & \multicolumn{4}{c}{Binary accuracy} & \multicolumn{4}{c}{Holdout retraining} \\
{} & Retain Acc & Forget Acc & Test Acc & MIA & Retain Acc & Forget Acc & Test Acc & MIA & Retain Acc & Forget Acc & Test Acc & MIA \\
\midrule
\midrule
Retrain & 99.995 ± 0.003 & 49.100 ± 1.910 & 60.699 ± 0.207 & 0.637 ± 0.013 & 99.996 ± 0.004 & 57.344 ± 0.956 & 60.585 ± 1.107 & 0.573 ± 0.019 & 99.980 ± 0.004 & 66.656 ± 1.832 & 60.072 ± 0.526 & 0.488 ± 0.015 \\
\midrule
Fine-tune RUM$^\mathcal{F}$ & 97.515 ± 9.221 & 57.233 ± 9.692 & 51.430 ± 6.930 & 0.477 ± 0.062 & 89.031 ± 2.309 & 54.056 ± 0.981 & 46.816 ± 2.642 & 0.492 ± 0.053 & 90.491 ± 2.924 & 67.911 ± 4.829 & 47.763 ± 0.255 & 0.368 ± 0.067 \\
Fine-tune shuffle & 86.404 ± 29.465 & 78.078 ± 42.146 & 47.103 ± 25.866 & 0.376 ± 0.598 & 88.963 ± 0.438 & 76.167 ± 1.363 & 47.436 ± 2.371 & 0.261 ± 0.065 & 80.539 ± 15.855 & 75.378 ± 11.318 & 45.989 ± 1.995 & 0.244 ± 0.117 \\
Fine-tune vanilla & 97.146 ± 11.969 & 79.356 ± 31.411 & 55.678 ± 12.037 & 0.301 ± 0.188 & 85.916 ± 0.795 & 52.889 ± 0.960 & 46.876 ± 0.977 & 0.476 ± 0.079 & 82.642 ± 1.333 & 63.178 ± 3.064 & 45.162 ± 1.034 & 0.363 ± 0.062 \\
\midrule
NegGrad+ RUM$^\mathcal{F}$ & 99.932 ± 0.080 & 56.678 ± 1.315 & 56.471 ± 2.238 & 0.585 ± 0.024 & 99.632 ± 0.152 & 71.078 ± 2.079 & 54.618 ± 0.935 & 0.334 ± 0.034 & 95.442 ± 2.226 & 73.078 ± 4.809 & 51.477 ± 3.137 & 0.281 ± 0.071 \\
NegGrad+ shuffle & 96.793 ± 0.110 & 83.778 ± 0.539 & 53.924 ± 1.024 & 0.217 ± 0.025 & 92.693 ± 19.577 & 87.722 ± 22.461 & 48.623 ± 14.360 & 0.177 ± 0.274 & 98.932 ± 2.600 & 95.678 ± 4.582 & 55.704 ± 5.997 & 0.083 ± 0.118 \\
NegGrad+ vanilla & 89.907 ± 1.114 & 53.167 ± 2.384 & 50.117 ± 1.047 & 0.500 ± 0.031 & 99.226 ± 2.797 & 92.244 ± 11.764 & 57.298 ± 7.970 & 0.156 ± 0.111 & 99.932 ± 0.132 & 94.444 ± 9.026 & 59.312 ± 1.005 & 0.069 ± 0.086 \\
\midrule
L1-sparse RUM$^\mathcal{F}$ & 90.010 ± 3.150 & 49.899 ± 2.632 & 46.896 ± 2.862 & 0.527 ± 0.024 & 96.847 ± 0.935 & 57.356 ± 2.136 & 52.504 ± 2.696 & 0.479 ± 0.043 & 96.108 ± 2.207 & 70.011 ± 2.402 & 51.990 ± 0.899 & 0.384 ± 0.022 \\
L1-sparse shuffle & 81.317 ± 8.350 & 65.978 ± 11.740 & 46.023 ± 4.367 & 0.360 ± 0.160 & 95.620 ± 3.593 & 82.967 ± 4.120 & 51.884 ± 2.405 & 0.235 ± 0.002 & 95.860 ± 1.253 & 87.778 ± 3.182 & 51.497 ± 0.366 & 0.168 ± 0.013 \\
L1-sparse vanilla & 87.214 ± 2.139 & 49.433 ± 1.168 & 47.343 ± 1.151 & 0.528 ± 0.016 & 83.734 ± 2.898 & 55.311 ± 1.265 & 48.710 ± 2.527 & 0.466 ± 0.019 & 85.327 ± 1.300 & 66.367 ± 1.547 & 48.030 ± 1.254 & 0.364 ± 0.040 \\
\midrule
SalUn RUM$^\mathcal{F}$ & 82.377 ± 2.398 & 54.622 ± 5.034 & 47.990 ± 1.540 & 0.474 ± 0.136 & 80.224 ± 3.982 & 58.333 ± 5.413 & 47.123 ± 3.210 & 0.431 ± 0.098 & 78.037 ± 6.344 & 69.900 ± 5.152 & 45.729 ± 2.459 & 0.239 ± 0.044 \\
SalUn shuffle & 84.977 ± 36.200 & 72.622 ± 43.082 & 50.430 ± 15.871 & 0.313 ± 0.380 & 74.747 ± 6.517 & 63.533 ± 4.898 & 44.256 ± 2.857 & 0.341 ± 0.093 & 76.539 ± 4.944 & 74.967 ± 14.223 & 45.456 ± 1.959 & 0.188 ± 0.043 \\
SalUn vanilla & 98.336 ± 0.254 & 85.522 ± 2.569 & 56.945 ± 1.357 & 0.322 ± 0.035 & 79.726 ± 4.278 & 66.567 ± 4.349 & 47.116 ± 2.527 & 0.402 ± 0.120 & 78.775 ± 9.791 & 81.244 ± 7.437 & 49.517 ± 2.461 & 0.201 ± 0.100 \\
\bottomrule
\end{tabular}}
\caption{Tiny-ImageNet with VGG-16}
\end{subtable}
\label{tab:acc-full}
\end{table}

\subsubsection{Stability analysis of proxies} \label{sec:proxy-stability}
\begin{figure}[t]
\centering
\begin{subfigure}[b]{0.32\textwidth}
 \centering
\includegraphics[scale=0.2]{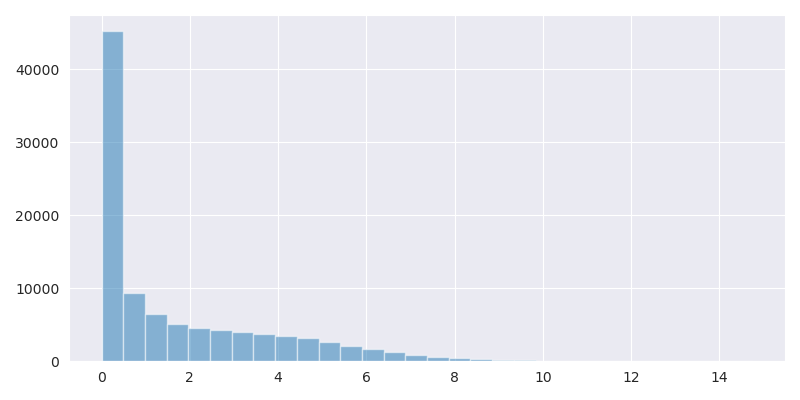}
 \caption{Step 0}
\end{subfigure}
%  \hfill
\begin{subfigure}[b]{0.32\textwidth}
 \centering
\includegraphics[scale=0.2]{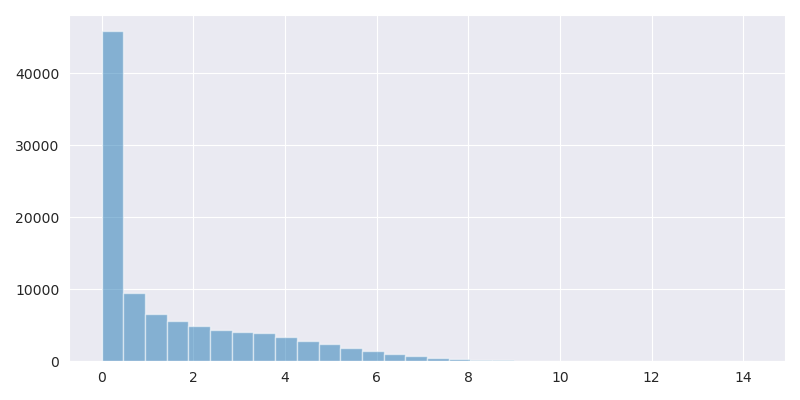}
 \caption{Step 1}
\end{subfigure}
\begin{subfigure}[b]{0.32\textwidth}
 \centering
\includegraphics[scale=0.2]{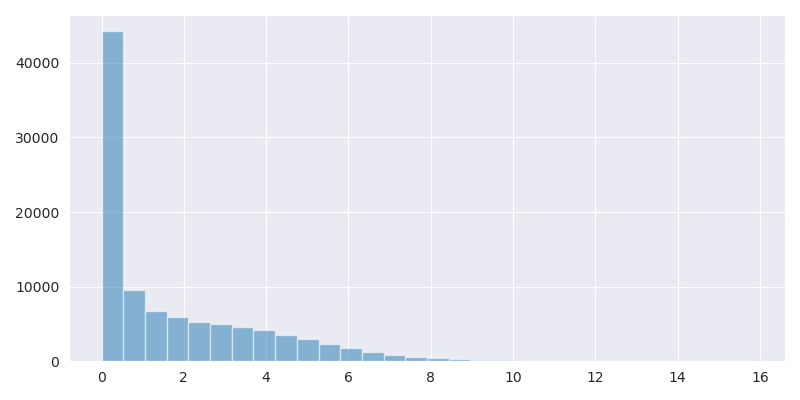}
 \caption{Step 2}
\end{subfigure}
\begin{subfigure}[b]{0.32\textwidth}
 \centering
\includegraphics[scale=0.2]{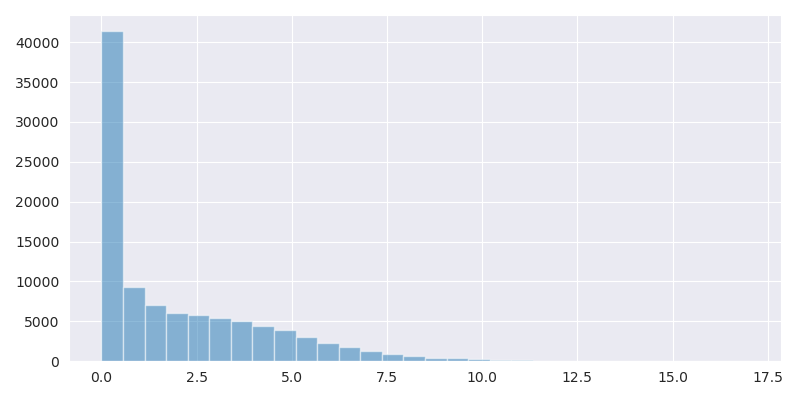}
 \caption{Step 3}
\end{subfigure}
\begin{subfigure}[b]{0.32\textwidth}
 \centering
\includegraphics[scale=0.2]{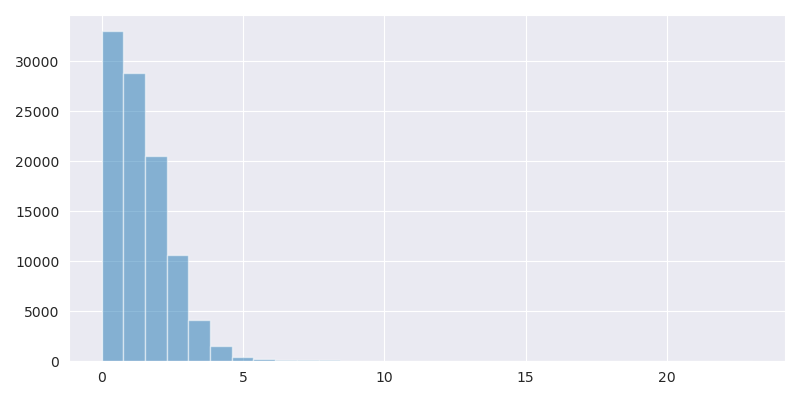}
 \caption{Step 4}
\end{subfigure}
\begin{subfigure}[b]{0.32\textwidth}
 \centering
\includegraphics[scale=0.2]{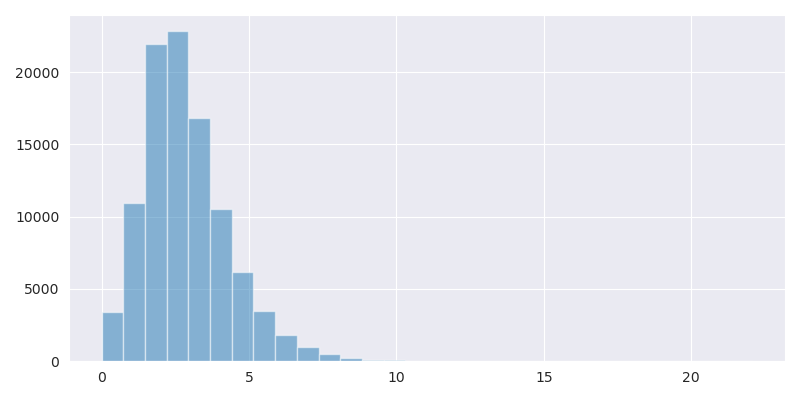}
 \caption{Step 5}
\end{subfigure}
\caption{Distribution of proxy values before and after each unlearning step, using holdout retraining as the proxy and NegGrad+ as the unlearning baseline with the vanilla approach, evaluated on Tiny-ImageNet with VGG-16 model architecture.}
\label{fig:proxy-distro-seq}
\end{figure}

We use NegGrad+ as a baseline and apply both RUM$^\mathcal{F}$ and vanilla as unlearning approaches for comparison. 
Each approach (vanilla or RUM$^\mathcal{F}$) is sequentially applied over 5 steps, and we track ToW, ToW-MIA, and runtime after each step. 
Table \ref{tab:seq-tow-towmia} presents the detailed results of the stability analysis discussed in Section \ref{sec:stability}, and Figure \ref{fig:proxy-distro-seq} visualizes the distribution of proxy values across the unlearning steps.

\begin{table}[t]
\centering
\caption{Unlearning performance (evaluated by ToW and ToW-MIA) and runtime across 5 sequential steps.
The unlearning algorithm $\mathcal{U}$ (NegGrad+) is applied at each step using two approaches: (i) "vanilla" and (ii) RUM$^\mathcal{F}$." Results are averaged over three runs and reported with 95\% confidence intervals.}
\begin{subtable}{\textwidth}
\centering
\resizebox{\textwidth}{!}{
\begin{tabular}{lccc|ccc|ccc}
\toprule
\cmidrule(r){1-10}
& \multicolumn{3}{c}{NegGrad+ RUM$^\mathcal{F}$} & \multicolumn{3}{c}{NegGrad+ vanilla} & \multicolumn{3}{c}{Retrain} \\
{} & ToW ($\uparrow$) & ToW-MIA ($\uparrow$) & Runtime (s) & ToW ($\uparrow$) & ToW-MIA ($\uparrow$) & Runtime (s) & ToW ($\uparrow$) & ToW-MIA ($\uparrow$) & Runtime (s) \\
\midrule
Step 1 & 0.901 ± 0.035 & 0.791 ± 0.046 & 372.202 & 0.771 ± 0.079 & 0.661 ± 0.047 & 140.775 & 1.000 ± 0.000 & 1.000 ± 0.000 & 378.786 \\
Step 2 & 0.883 ± 0.029 & 0.737 ± 0.082 & 341.856 & 0.825 ± 0.031 & 0.691 ± 0.073 & 124.151 & 1.000 ± 0.000 & 1.000 ± 0.000 & 374.820 \\
Step 3 & 0.888 ± 0.059 & 0.771 ± 0.070 & 311.569 & 0.880 ± 0.047 & 0.743 ± 0.070 & 116.526 & 1.000 ± 0.000 & 1.000 ± 0.000 & 366.586 \\
Step 4 & 0.887 ± 0.047 & 0.740 ± 0.047 & 291.224 & 0.880 ± 0.041 & 0.729 ± 0.033 & 102.648 & 1.000 ± 0.000 & 1.000 ± 0.000 & 348.416 \\
Step 5 & 0.893 ± 0.081 & 0.748 ± 0.051 & 270.743 & 0.890 ± 0.025 & 0.743 ± 0.043 & 98.810 & 1.000 ± 0.000 & 1.000 ± 0.000 & 317.038 \\
\bottomrule
\end{tabular}}
\caption{CIFAR-10 with ResNet-18}
\end{subtable}

\begin{subtable}{\textwidth}
\centering
\resizebox{\textwidth}{!}{
\begin{tabular}{lccc|ccc|ccc}
\toprule
\cmidrule(r){1-10}
& \multicolumn{3}{c}{NegGrad+ RUM$^\mathcal{F}$} & \multicolumn{3}{c}{NegGrad+ vanilla} & \multicolumn{3}{c}{Retrain} \\
{} & ToW ($\uparrow$) & ToW-MIA ($\uparrow$) & Runtime (s) & ToW ($\uparrow$) & ToW-MIA ($\uparrow$) & Runtime (s) & ToW ($\uparrow$) & ToW-MIA ($\uparrow$) & Runtime (s) \\
\midrule
Step 1 & 0.816 ± 0.020 & 0.692 ± 0.030 & 846.290 & 0.716 ± 0.069 & 0.576 ± 0.069 & 483.439 & 1.000 ± 0.000 & 1.000 ± 0.000 & 5544.903 \\
Step 2 & 0.854 ± 0.081 & 0.776 ± 0.081 & 819.794 & 0.724 ± 0.028 & 0.634 ± 0.081 & 457.230 & 1.000 ± 0.000 & 1.000 ± 0.000 & 4874.518 \\
Step 3 & 0.808 ± 0.059 & 0.751 ± 0.038 & 800.202 & 0.729 ± 0.024 & 0.592 ± 0.041 & 448.514 & 1.000 ± 0.000 & 1.000 ± 0.000 & 4612.383 \\
Step 4 & 0.738 ± 0.043 & 0.723 ± 0.067 & 783.190 & 0.728 ± 0.040 & 0.674 ± 0.076 & 432.879 & 1.000 ± 0.000 & 1.000 ± 0.000 & 4548.031 \\
Step 5 & 0.652 ± 0.054 & 0.576 ± 0.047 & 755.287 & 0.668 ± 0.094 & 0.672 ± 0.058 & 420.158 & 1.000 ± 0.000 & 1.000 ± 0.000 & 4321.621 \\
\bottomrule
\end{tabular}}
\caption{Tiny-ImageNet with VGG-16}
\end{subtable}
\label{tab:seq-tow-towmia}
\end{table}

\end{document}